\documentclass{article}

\usepackage{arxiv}

\usepackage[utf8]{inputenc} 
\usepackage[T1]{fontenc}    
\usepackage{hyperref}       
\usepackage{url}            
\usepackage{booktabs}       
\usepackage{amsfonts}       
\usepackage{nicefrac}       
\usepackage{microtype}      
\usepackage{graphicx}
\graphicspath{ {./images/} }

\usepackage{caption}
\usepackage{subcaption}
\usepackage{longtable}
 \usepackage{amsmath}
\newcounter{customcounter}
\title{TreeNet: Layered Decision Ensembles}

\author{
 Zeshan Khan \\
 Department of Software Engineering\\
  FAST School of Computing\\
  National University of Computer and Emerging Sciences\\
  Islamabad, Pakistan \\
  \texttt{zeshan.khan@nu.edu.pk} \\
}

\begin{document}
\maketitle
\begin{abstract}
Within the domain of medical image analysis, three distinct methodologies have demonstrated commendable accuracy: Neural Networks, Decision Trees, and Ensemble-Based Learning Algorithms, particularly in the specialized context of genstro institutional track abnormalities detection. These approaches exhibit efficacy in disease detection scenarios where a substantial volume of data is available. However, the prevalent challenge in medical image analysis pertains to limited data availability and data confidence.
This paper introduces TreeNet, a novel layered decision ensemble learning methodology tailored for medical image analysis. Constructed by integrating pivotal features from neural networks, ensemble learning, and tree-based decision models, TreeNet emerges as a potent and adaptable model capable of delivering superior performance across diverse and intricate machine learning tasks. Furthermore, its interpretability and insightful decision-making process enhance its applicability in complex medical scenarios.
Evaluation of the proposed approach encompasses key metrics including Accuracy, Precision, Recall, and training and evaluation time. The methodology resulted in an F1-score of up to 0.85 when using the complete training data, with an F1-score of 0.77 when utilizing 50\% of the training data. This shows a reduction of F1-score of 0.08 while in the reduction of 50\% of the training data and training time. The evaluation of the methodology resulted in the 32 Frame per Second which is usable for the realtime applications. This comprehensive assessment underscores the efficiency and usability of TreeNet in the demanding landscape of medical image analysis specially in the realtime analysis.
\end{abstract}

\keywords{TreeNet \and Medical Image Analysis \and Decision Trees}

\section{Introduction}
\label{sec:intro}
The evolving field of medical image analysis, enabling the extraction of necessary information from medical image modalities of computed tomography (CT), magnetic resonance imaging (MRI) \cite{khoo1997magnetic},  X-rays \cite{hiriyannaiah1997x}, and ultrasound \cite{dhawan2011medical}, etc. The field of image analysis has received significant advancement driven by convergence and advances in the domains of machine learning and deep learning with imaging techniques. The development of sophisticated algorithms and techniques to automate the analysis and interpretation of medical images, facilitating more accurate diagnoses and treatment planning. Notable progress has been made in the segmentation of anatomical structures, detection of abnormalities, and quantitative assessment of disease progression.

One seminal contribution to the field of medical image analysis is the work on deep learning-based methods, particularly convolutional neural networks (CNNs). Convolutional neural networks have demonstrated remarkable success in tasks such as image classification, segmentation, and object detection \cite{krizhevsky2012imagenet,he2016deep,khan2018majority,khan2021medical,khan2021medico}.

Moreover, advancements in medical image analysis have not only focused on algorithmic improvements but also on the integration of multimodal data for a comprehensive understanding of complex medical conditions \cite{aine2017multimodal}.

The recent advancements in the machine learning and deep learning is generating a great results in almost every learning domain. The most used architectures in the recent years for the general machine learning and specialized for image classification are as follows:
\begin{description}
    \item [Neural Networks:] Neural Network architectures possess several distinctive characteristics that define their behavior and performance. Firstly, they operate in a layered processing manner, where information flows through successive layers of nodes, allowing for hierarchical feature representation. In each layer, feature transformations take place, enabling the network to learn complex patterns and representations from the data. However, this complexity comes with a trade-off, as neural networks require a sufficient model complexity, often determined by a higher number of hyperparameters, which influence their performance significantly. Consequently, the success of a neural network heavily relies on careful hyperparameter tuning.
    
    Another important aspect of neural networks is their susceptibility to the gradient descent problem, which may lead to convergence issues during training. Analyzing the theoretical aspects of neural networks can be challenging, particularly because hyperparameters profoundly impact the model structure and behavior. Additionally, neural networks demand a substantial amount of data for effective training, as their model complexity is data-dependent, and deep models may outperform shallower ones when provided with abundant data \cite{ayidzoe2023sinocaps,jha2021comprehensive,ramzan2023gastrointestinal,zhou2018unet++,jha2020doubleu,hicks2018deep}.
    
    Despite their strengths, neural networks are not always superior to other models. In specific cases, models like XGBoost and Random Forest have shown comparable or better performance, as indicated by Tseng et al. \cite{tseng2023optimized}. To overcome some of the complexities and potential performance issues, techniques like skip connections, pruning, and quantization have been employed, which may mitigate the impact of model complexity on performance.
    
    \item [Ensemble Learning:] Ensemble Learning architectures possess distinct characteristics that set them apart from single-model approaches. They offer several advantages, including improved generalization compared to using a single model. The ensemble learning can reduce overfitting to enhance predictive performance on unseen data by using impact of multiple models.
    
    The performance of the ensemble learning depends on the prediction of individual models and the diversity of the individual models within the ensemble. Each model should capture the unique perspectives on the data, to contribute to a more robust and accurate collective prediction. The diversity can be achieved in ensemble learning by data manipulation, feature manipulations and learning parameter manipulation. Ensemble learning aggregates the predictions of individual models through techniques of averaging and  weighted averaging where each model's output contributes to the final prediction based on its performance or reliability \cite{ramzan2023gastrointestinal,jha2021comprehensive}. This integration of diverse models yield an ensemble that can outperform individual models and boost up overall predictive performance.

    \item [Decision Trees:] Tree-based methodologies have inherited advantages of interpretability, flexibility, and feature importance based decisions. These behavioral aspects of the decision trees makes them imperative deciders for the explainable and understandable. These properties of the decision trees are proved to be effective in the cases of limited data availability \cite{wen2021machine,zhou2014capsule,choe2023machine,zhang2023machine}. The interpretability and discernible decision pathways facilitated by these tree-based approaches significantly contribute to their efficacy in medical imaging scenarios, enhancing the transparency and interpretative insights crucial for clinical applications.
    Tree based approaches may face some of the challenges of overfitting due to higher depth of the trees with respect to data availability. Another challenge in the decision trees is about the expressiveness which may cause decision trees lesser effective in case of complex relationships in the data. Despite these challenges, understanding the strengths and limitations of tree-based approaches is essential for effectively using them in various machine learning tasks \cite{tseng2023optimized}.
\end{description}

The study investigates various types of models for the classification of medical image and proposed a hybrid model named Layered Decision Ensemble (TreeNet), inspired by Deep Forest \cite{zhou2019deep}. TreeNet is a hybrid of neural network as feed forward, decision trees and random forest classifier.
TreeNet is a layer-by-layer processing like forward pass of the residual neural network with the decision trees ensembles in each layer which is a behavior of tree forest. 

The evaluation framework employed in this study was designed to capture multiple dimensions of model performance rather than relying on a single metric. Particular emphasis was placed on detection speed, predictive accuracy, data efficiency, and both training and inference times. These aspects were selected not only for their technical importance but also for their relevance to real-world deployment, where computational resources and time constraints often define feasibility. By aligning the assessment with these practical considerations, the proposed model was positioned to demonstrate strengths in efficiency and adaptability while maintaining competitive accuracy.

The proposed methodology has been rigorously applied to benchmark datasets, namely Kvasir V1, Kvasir V2, and Hyper Kvasir \cite{riegler2017multimedia,pogorelov2017kvasir,pogorelov2018medico,borgli2020hyperkvasir}. The outcomes of this approach demonstrate a commendable equilibrium between precision and recall, yielding an improved F1 score compared to certain preceding studies. Notably, the approach achieves remarkable efficiency with minimal training time and computational demands, thereby contributing to reduced carbon emissions. The F1-scores obtained on the benchmark datasets of Kvasir V1, Kvasir V2, and Hyper Kvasir stand at 0.75, 0.78, and 0.82, respectively. A pivotal aspect of this approach is its notable reduction in training and inference times. For instance, the training duration on Kvasir V1 is under 40 minutes, even on a CPU-only machine devoid of a Graphic Processing Computational Unit. Furthermore, the approach attains an impressive 30 frames per second, exemplifying its efficacy in real-time applications.

The subsequent sections of this paper are meticulously organized to offer a comprehensive exploration of the research. Section ~\ref{sec:work} critically examines the related work, shedding light on prior research in the domain. The innovative methodology proposed is expounded upon in Section ~\ref{sec:approach}, providing a lucid and detailed elucidation of the novel approach employed. Section ~\ref{sec:datasets} meticulously details the datasets utilized in the research, furnishing essential context for the experiments conducted in the subsequent sections.

Section ~\ref{sec:eval} presents the experimental set-up, detailing the procedures followed during the evaluation. The results of the experiments are discussed in Section ~\ref{sec:results}, offering an analysis of the model's performance. Furthermore, Section ~\ref{sec:discussion} provides additional insights and discussions regarding the obtained results. Section ~\ref{sec:conclusion} serves as the conclusion, summarizing the key findings of the study. Lastly, Section ~\ref{sec:future} outlines the future work, suggesting potential directions for further research and development.

\section{Related Work}
\label{sec:work}
In the realm of medical image analysis, notable advancements have been attained through the implementation of three predominant machine learning approaches. Firstly, Artificial Neural Networks (ANNs) have emerged as powerful models, demonstrating remarkable success in tasks such as image classification and segmentation \cite{jha2021comprehensive,ali2020depth,siddiqui2021attention,khan2022voting}. ANNs, inspired by the human brain's neural architecture, excel in learning intricate patterns and relationships within medical images, contributing to enhanced diagnostic accuracy and efficiency.

Ensemble learning has shown promise in medical imaging by combining multiple models to improve accuracy and reduce bias. This collective strength often leads to more reliable outcomes than single models. Yet, the gains come at a cost. Training and deploying ensembles can be resource-intensive, slowing down analysis where speed is critical. Their complexity also makes them harder to interpret, raising questions of trust when clinicians need clear justifications for decisions \cite{jha2021comprehensive,jha2025validating,khan2022voting,khan2021medico,khan2023real,khan2018majority}.

Tree-based models, in contrast, are valued for their simplicity and interpretability. Their step-by-step logic can be traced, making predictions easier to understand and validate. But this transparency has limits. Decision trees are prone to overfitting, struggle with noisy data, and often lack the depth to capture subtle patterns in complex images. While useful for clarity, they rarely achieve the robustness required for clinical deployment on their own \cite{wen2021machine,zhou2014capsule,choe2023machine,zhang2023machine,jha2021comprehensive,jha2025validating}.

\subsection{Artificial Neural Networks based Models}
Artificial Neural Networks (ANNs) have reshaped medical image analysis by capturing complex, non-linear patterns that traditional models often miss \cite{shen2017deep,mall2023comprehensive,zhang2023applying}. Built from layered nodes, they progressively extract meaningful features, turning raw images into clinically useful insights. Among them, Convolutional Neural Networks (CNNs) stand out, achieving remarkable performance in recognition and segmentation tasks \cite{xia2024comprehensive,khan2021medical,jha2025validating,khan2021medico,bertels2022convolutional,khan2023real}. Their layered learning uncovers subtle textures and hidden structures, making them powerful tools for detecting anomalies that might escape human eyes. Moreover, the adaptability of ANNs allows them to continuously learn and improve their performance with additional data, reinforcing their efficacy in evolving medical scenarios \cite{kumari2023continual}.

\begin{equation}
\label{eq:nn}
    h_i = \sigma(W_i^T x + b_i)
\end{equation}

\begin{equation}
\label{eq:nnerror}
L(y, \hat{y}) = \frac{1}{N} \sum_{i=1}^N (y_i - \hat{y}_i)^2
\end{equation}

\begin{equation}
\label{eq:nnbprob}
    w_{new} = w_{old} - \eta \cdot 
\begin{bmatrix}
\frac{\partial L}{\partial w_1} \\
\frac{\partial L}{\partial w_2} \\
\vdots \\
\frac{\partial L}{\partial w_n}
\end{bmatrix}
\end{equation}

Considering a Neural network with L layers and having N neurons in each layer shown in Equation~\ref{eq:nn} and the back propagation in Equation~\ref{eq:nnerror}. The training of the network is performed using a dataset of size D for training and batch size B. The network will need a matrix multiplication and addition of bias proceeded by application of activation function as shown in Equation~\ref{eq:nnbprob}.

The inference time of the network can be computed from number of layers, Training factors and number of neurons in each layer. Equation~\ref{eq:nnttime} shows the training time of a network considering N neurons in each layer of the network.
\begin{equation}
\label{eq:nnitime}
Inference-Time =  L \times (N^2 + 2N)
\end{equation}

While training time reflects the computational effort required to optimize the network's weights based on input data and associated labels, inference time focuses on the efficiency of the trained model when making predictions on unseen data. Both aspects are critical in evaluating the overall performance and deployment feasibility of neural networks, especially in real-time or resource-constrained environments.

The inference time of the network can be computed from number of layers and number of neurons in each layer. Equation~\ref{eq:nnitime} shows the inference time of a network considering N neurons in each layer of the network.

The training time of the network can be computed from number of layers, Training factors and number of neurons in each layer. Equation~\ref{eq:nnttime} shows the training time of a network considering N neurons in each layer of the network.

\begin{equation}
\label{eq:nnttime}
Training-Time= e \times (D / B) * (2L \times (N^2 + 2N) + L \times N^2)
\end{equation}

To evaluate the real-world applicability of neural networks, it is essential not only to understand their theoretical inference time but also to examine how different architectures perform in practical scenarios. Among various types of neural networks, Recurrent Neural Networks (RNNs) and their variants have shown considerable success in handling sequential and temporal data, particularly in the domain of medical diagnostics \cite{ozturk2021residual}.

Sushovan Chaudhury and Kartik Sau used the Gated Recurrent Units (GRUs) a sub class of Recurrent Neural Networks (RNNs) called GRU-RNN \cite{chaudhury2023blockchain}. The GRU-RNN when applied for the identification of brest cancer in the locally collected dataset and resulted in the accuracy of 0.97 with the recall of 0.95 and f1-score of 0.95.

In the paper "Transfer Learning with Prioritized Classification and Training Dataset Equalization for Medical Objects Detection" by Olga Ostroukhova et al., the authors introduce a novel approach developed by the SIMULA organizer team for the MediaEval 2018 Multimedia for Medicine: the Medico Task \cite{ostroukhova2018transfer}. Leveraging transfer-learning-based image classification techniques, the study emphasizes the simplicity of implementing multi-class image classifiers and explores avenues to enhance classification performance without necessitating extensive modifications to deep neural network architectures. The primary objective is twofold: to establish a baseline for the Medico task and to demonstrate the efficacy of out-of-the-box classifiers within the context of medical image analysis.

Another research evaluated on the dataset of MediaEval 2018 \cite{pogorelov2018medico} is based on Weighted Discriminant Embedding (WDE), which projects data into low dimensional feature space. The approach resulted in a lower detection results of F1-score of 0.48 with a speedy detection.

The 2018 Medico Multimedia Task \cite{pogorelov2018medico} Submission by Team NOAT presents an a tree stage methodology for the classification of medical images portraying the human gastrointestinal tract \cite{steiner20182018}. The approach starts with deep features extraction from each medical image employing pre-trained deep-learning models. Subsequently, LIRE is employed for feature indexing in the second stage, facilitating efficient retrieval of similar images \cite{lux2008lire}. In the final stage, predictions are made based on the search results. The team's approach demonstrates notable success, achieving a Matthews Correlation Coefficient (MCC) score of 0.54 and an accuracy of 0.94, coupled with F1-scores ranging from 0.36 to 0.57 across diverse methodologies.

The cited research highlights the distinctive characteristics of Neural Network-based architectures compared to ensemble learning and tree-based learning approaches. The characteristics identified are as follows:

\subsubsection{Feature Transformation due to Layered Processing} 
Neural networks excel due to their ability to process data through multiple layers of interconnected nodes, allowing them to progressively transform raw data into meaningful, higher-level features \cite{taherdoost2023deep,jogin2018feature}. Each layer in the network builds upon the output of the previous layer, extracting increasingly complex patterns and relationships from the data. This layered structure is crucial for effective feature transformation, as it enables the network to learn and refine features at each stage \cite{shen2017deep}. The network's performance improves significantly with the availability of large, high-quality training datasets, highlighting the importance of data in optimizing feature transformation \cite{chen2022recent}. By leveraging this layered processing, neural networks are able to transform features in a way that enhances their decision-making capabilities, making them powerful tools for a wide range of tasks.

\subsubsection{Model Complexity}
Neural networks exhibit a notable proficiency in effectively managing intricate and high-dimensional datasets, thereby establishing them as eminently suitable for tasks demanding sophisticated modeling. This inherent capability underscores the efficacy of neural networks in addressing challenges associated with complex data structures, thereby enhancing their applicability across a spectrum of tasks requiring advanced model intricacy. The layered architecture inherent in neural networks, marked by multiple perceptions and activation functions, significantly contributes to the inherent complexity of the model. This complexity, in turn, renders the model challenging to comprehend and theoretically justify with regard to result accuracy. The time complexity of a neural network with $L$ layers and each layer having $n_l$ neurons can be expressed with the forward pass and backward pass. The Forward pass takes the time shown in Equation~\ref{eq:forwardpass} while the backword time is double to the forward for each epoch. The total time for $E$ epochs is shown in Equation~\ref{eq:dltime}.
\begin{equation}
\label{eq:forwardpass}
O\left( \sum_{l=1}^{L} n_{l-1} \times n_l \right)
\end{equation}

\begin{equation}
\label{eq:dltime}
O\left( 2E \times \sum_{l=1}^{L} n_{l-1} \times n_l \right)
\end{equation}

\subsubsection{Hyper-parameters}
Hyper-parameters critically shape a neural network’s learning capacity and generalization. Subtle choices in learning rate, batch size, and depth can drastically alter performance in capturing the intricate patterns of medical images \cite{liao2022empirical,koutsoukas2017deep,smithson2016neural}.

\subsubsection{Differential Mode (Weight Adjustment)}
Weight adaptation through gradient-driven optimization enables networks to iteratively refine internal representations. This process lies at the core of learning, allowing models to align with data distributions and uncover hidden structures in complex medical imagery \cite{ruder2016overview}.

\subsubsection{Gradient Descent Challenges}
Gradient Descent, while a fundamental optimization algorithm for training neural networks, is not without its challenges. One notable issue is the potential entrapment in local minima during the training process \cite{liu2021conflict,chatterjee2022convergence,arora2018convergence}. This phenomenon arises when the algorithm converges to a suboptimal solution instead of the global minimum, hindering the network's ability to reach the most optimal configuration. Consequently, researchers and practitioners in the field of machine learning must navigate this challenge strategically, employing various techniques such as adaptive learning rates or stochastic variations of the algorithm to mitigate the risk of stagnation in suboptimal states. Gradient descent is sensitive to the learning rate, computationally costly, and highly dependent on initial conditions \cite{glorot2010understanding,bengio2000gradient,ruder2016overview}.

\subsubsection{Complexity-Performance Relationship}
The intricate interplay between model complexity and performance constitutes a pivotal consideration in the meticulous design of neural networks. This fundamental relationship underscores the delicate balance required to achieve optimal performance without succumbing to the pitfalls of excessive complexity. While a more complex model may possess the capacity to capture intricate patterns within the data, it concurrently introduces the risk of overfitting, where the model becomes excessively attuned to the training data at the expense of generalizability to unseen data \cite{freire2021performance,lee2020neural,boccato2024beyond}. The historically neural networks from LeNet-1 to the Transformers, the network is getting more complex in terms of layers and getting higher performance \cite{lecun1998gradient,lin2022survey}

\subsection{Ensemble Learning Based Models}

Ensemble Learning, characterized by the combination of multiple models to enhance overall performance, has proven highly effective as it reduces the biasesness \cite{dong2020survey,dasarathy1979composite,kearns1994learning,schapire1990strength}. Techniques such as bagging and boosting involve training diverse models and aggregating their predictions to achieve a more robust and accurate outcome \cite{breiman1996bagging,domingos1997does}. In medical diagnostics, where the consequences of misclassification can be critical, ensembles contribute to improved reliability by reducing the risk of individual model errors. Random Forests, a popular ensemble method based on decision trees, have been successful in handling complex medical data by capturing diverse aspects of the information \cite{breiman2001random}. The collaborative nature of ensemble learning not only enhances predictive accuracy but also fosters resilience against noisy or incomplete data, a common challenge in medical imaging applications. The identified characteristics are as follows:
\subsubsection{Generalization}
Generalization reflects a model’s ability to make reliable predictions on unseen data, a property crucial in medical imaging where test cases often differ from training conditions. Ensemble learning strengthens this ability by combining multiple learners, reducing overfitting, and capturing broader data representations. As a result, ensembles frequently outperform single models in handling the complexity of medical images \cite{ganaie2022ensemble,sagi2018ensemble,wolpert1992stacked}.

\subsubsection{Diversity-driven Performance}
The strength of ensembles lies in diversity. Different models capture different aspects of the same data, and their integration reduces systematic errors. This diversity acts as a safeguard against the weaknesses of individual models, leading to more stable and context-aware predictions \cite{wood2023unified,minku2009impact}.

\subsubsection{Input Feature Manipulation}
Introducing diversity at the feature level enriches ensemble learning further. By training models on different subsets of features, each learner becomes sensitive to unique structures or patterns. This specialization allows the ensemble to better capture the complexity of medical images and to remain robust under varied imaging conditions \cite{breiman2001random}.

\subsubsection{Output Manipulation}
At the output stage, combining predictions through majority voting, averaging, or weighted fusion helps stabilize decisions and minimize individual model variability. This integration creates a consensus that is often more accurate, reliable, and trustworthy than any single prediction \cite{sagi2018ensemble,dietterich2000ensemble}.

\subsection{Decision Trees Based Models}
Decision Trees bring clarity to medical image analysis, offering a simple yet powerful way to trace how decisions unfold. By splitting data through binary choices, they create a transparent path from input to diagnosis which is explainable for healthcare practitioners. This interpretability builds trust, a quality essential for responsible adoption of AI in clinical workflows. They also highlight the exact image features driving a diagnosis, making them invaluable when explanations matter as much as predictions. Their simplicity and clarity make them suitable for collaborative decision-making between machines and healthcare professionals, as the logic behind the model's predictions can be easily communicated and validated. Yet, their simplicity can backfire: Decision Trees often overfit, struggle with complex patterns, and may fail to scale when data grows richer and noisier.

Cheng-Jui Tseng and Changjiang Tang proposed an optimized eXtreme Gradient Boosting (PSO-XGBoost) approach \cite{tseng2023optimized}. The approach is using XGBoost for classification of the brain tumor with Particle Swarm Optimization (PSO) to choose characteristics. The model is evaluated on a brain tumer detection dataset of Kaggle \cite{kaggle-brain} and achieved 0.97 accuracy, 0.97 specificity, 0.98 precision, and 0.98 recall.

A feature engineering approach with various classifiers is applied for the Lung and colon cancer classification \cite{hage2022lung}. The XGBoost out performed when the models of XGBoost, SVM, RF, LDA, MLP and LightGBM evaluated on the dataset of LC25000 \cite{borkowski2019lung}. The XGBoost resulted in the precision, recall, accuracy and f1-score of 0.99.

Pulmonary Embolism Cases in Chest CT Scans are classified using VGG16 features on XGBoost for classification \cite{dua2022classifying}. The approach is tested on the dataset of Pulmonary Embolism CT Dataset \cite{colak2021rsna} and resulted in the accuracy of 0.98 and a sensitivity of 0.97.

SMOTE-RFE-XGBoost a model with the class imbalance resolution with SMOTE, features selection using recursive feature elimination (RFE) with the XGBoost classification, was applied to a binary class dataset \cite{zhang2023classification}. The model resulted in the accuracy and F1 values of 0.98 and 0.87 respectively.

ATLAAS is an automatic decision tree-based learning algorithm for image segmentation in positron emission tomography (PET) \cite{berthon2016atlaas}. The algorithm is applied on a dataset of 100 PET scans with identified contour. The ATLAAS, when applied on the dataset, resulted in the accuracy of 0.93 and Dice similarity coefficient (DSC) of 0.88.

The research mentioned above highlights various characteristics of tree-based architectures when compared to neural networks and ensemble learning. Tree-based architectures, such as decision trees, exhibit specific traits that differentiate them from other learning approaches.

In contrast to neural networks, which involve layered processing and complex feature transformations, tree-based architectures employ a hierarchical, tree-like structure for decision-making. Each node in the tree represents a decision based on a specific feature, making them more interpretable and understandable. Additionally, tree-based models are non-parametric, meaning they do not assume any specific data distribution, allowing them to be more flexible in handling different types of data.

Compared to ensemble learning methods, where multiple models are combined for improved performance, tree-based architectures can provide feature importance information, indicating which features have the most significant impact on predictions. They can also handle missing values and categorical features more naturally.

Moreover, tree-based models exhibit certain limitations. They can be prone to overfitting, especially when the trees are deep and not pruned effectively. Additionally, decision trees may struggle with complex relationships in the data, requiring more advanced techniques to capture intricate patterns.

The subsequent points elucidate characteristics emblematic of tree-based learning:

\subsubsection{Tree-based Decisions}
A decision tree is a structured representation that simplifies a complex decision-making process into a series of elementary choices. These choices are made by evaluating feature values at each node and then proceeding through the branches until reaching a leaf node. The final decision or prediction is determined by the characteristics of the leaf node.

\subsubsection{Feature Priority Organization}
When constructing a decision tree, the algorithm selects the most suitable feature to divide the data at each node. The features are organized based on their capacity to effectively distinguish the data and create distinct subsets, ultimately leading to improved decision-making \cite{grabczewski2005feature,zhou2021feature,sugumaran2007feature}.
\subsubsection{Time-Optimal Feature Ignorance}
Decision trees can reduce computational burden by ignoring irrelevant features at specific nodes, lowering complexity without compromising predictive focus \cite{grabczewski2005feature,zhou2021feature}.

\subsubsection{Interpretability}
Their transparent, rule-based structure enables clear tracing of decisions, a critical asset in clinical workflows where explainability builds trust between data scientists and healthcare practitioners \cite{gilpin2018explaining,sagi2021approximating,silva2020optimization}.

\subsubsection{Non-parametric}
Free from distributional assumptions, decision trees adapt flexibly to heterogeneous data, making them suitable for varied medical imaging tasks.

\subsubsection{Feature Importance based Decision}
By ranking features based on discriminative power, trees highlight clinically relevant attributes, offering not only predictions but also interpretable insights into diagnostic cues \cite{grabczewski2005feature,sugumaran2007feature}.

\subsubsection{Handling Non-linear Relationships}
Decision trees naturally capture non-linear relationships through recursive partitioning, enabling them to model complex dependencies often present in medical data.

\subsubsection{Limitations}
However, decision trees are prone to overfitting, especially in deep structures without pruning, and can be unstable, where minor data perturbations alter the entire tree \cite{dietterich2000ensemble}. Their greedy splitting often leads to suboptimal structures, while discretization may oversimplify continuous features. They also risk bias towards dominant classes in imbalanced datasets, limiting sensitivity to rare but clinically important cases.

\subsubsection{Expressive Boundaries}
Their hierarchical framework struggles with higher-order interactions and intricate dependencies, reducing accuracy in complex imaging domains. Ensemble extensions like Random Forests partly alleviate these shortcomings \cite{sagi2021approximating}.

\subsubsection{Additional Strengths}
Decision trees naturally handle missing values, show robustness to outliers, and integrate categorical features without extensive preprocessing, reinforcing their practicality in real-world datasets.

Some of the features of decision trees are not necessary for the dataset of images, but are necessary in some other tasks.
\begin{description}
\label{best-features}
  \item[Handling Missing Values:] One of the strengths of decision trees is their ability to work with missing values. Instead of requiring every gap in the data to be filled, the tree can still make decisions using the information that is available. This makes them practical for real-world datasets, where missing entries are common and filling them in can sometimes create bias.
  \item[Robust to Outliers:] Decision trees are also less sensitive to outliers compared to many other models. Since the data is split into smaller groups, an outlier usually affects only a specific branch rather than the whole model. This makes decision trees a good choice for datasets that might contain unusual values, though very extreme cases can still have some impact.
  \item[Easy Handling of Categorical Features:] Decision trees offer a notable advantage in their innate ability to seamlessly handle categorical features without requiring the common preprocessing step of one-hot encoding. Unlike some machine learning models that necessitate the transformation of categorical variables into binary indicators, decision trees can directly incorporate categorical attributes into their splitting criteria. The tree-building process involves selecting optimal categorical splits, making decision trees well-suited for datasets containing a mix of categorical and numerical features. This inherent capability simplifies the modeling process and contributes to the interpretability of the tree structure, as the branches can directly represent decisions based on categorical attribute values. The ease of handling categorical features without the need for extensive preprocessing enhances the practicality and efficiency of decision trees, particularly in scenarios where categorical variables play a significant role in the predictive task.
\end{description}

\subsection{Comparison of the Characteristics}
\label{comparisn_literature}
Recent literature presents a diverse set of strategies across neural networks, ensemble learning, and decision tree-based approaches, each offering specific strengths and limitations. A critical review highlights several recurring characteristics across recent works that influence model effectiveness in terms of training efficiency, inference time, generalization ability, and interpretability.
\subsubsection{Feature Transformation} Feature transformation has emerged as a key mechanism in enhancing model performance. Ying et al. (2024) introduced NEAT, a graph-contrastive framework that enables deep unsupervised feature transformation by capturing non-linear feature interactions, which contributes to improved generalization in downstream tasks \cite{ying2024unsupervised}. Similarly, MOAT, proposed by Wang et al. (2024), uses reinforcement learning to guide autoregressive transformation of features, significantly boosting classification accuracy across various domains \cite{wang2025towards}.
    \subsubsection{Reduced Architectural Complexity}
In efforts to reduce model complexity, TabPFNv2 (2023) demonstrates that simplified architectures, trained on synthetic tabular distributions, can match or exceed the performance of traditional deep learning models while maintaining a fraction of the complexity \cite{xiao2024traceable}. This is particularly important in scenarios where computational resources are limited, and model interpretability is preferred.
   \subsubsection{Gradient-Free Optimization}
Avoidance of gradient descent in recent models, particularly foundation models like TabPFNv2, is another promising trend. These models eliminate the need for per-task optimization by employing a transformer trained across a diverse distribution of tasks, enabling near-instant inference without retraining \cite{xiao2024traceable}.
    \subsubsection{Generalization}
Generalization is further supported through ensemble learning and unsupervised transformations. Recent frameworks such as MOAT and NEAT offer task-agnostic improvements, where diverse representations and transformations allow the models to perform robustly even with minimal supervision \cite{ying2024unsupervised,wang2025towards}.
    \subsubsection{ Feature Prioritization}
The prioritization of relevant features is also addressed in newer models. SMART (2023) and TabPFNv2 include mechanisms for feature prioritization, enabling the model to learn which features are most influential for predictions while discarding irrelevant information \cite{xiao2024traceable}.
    \subsubsection{ Interpretability}
The importance of interpretability has also gained attention, especially in fields requiring explainable decision-making. Methods such as traceable self-optimizing transformations enable a clearer understanding of how features are manipulated and how decisions are made \cite{xiao2024traceable}. These mechanisms help bridge the gap between model accuracy and transparency.
    \subsubsection{ Un-Bias Towards Dominant Classes}
The issue of bias toward dominant classes is acknowledged in recent fairness-aware frameworks. For example, König et al. (2024) emphasize using explainability tools to identify and mitigate bias in imbalanced datasets, ensuring models remain fair and generalizable across minority classes \cite{gutierrez2024study}.

In the realm of medical image classification, where the challenges of limited data availability and the need for rapid training and inference outcomes take precedence, three distinct approaches demonstrate valuable attributes. The analysis of these approaches is systematically presented in Table~\ref{tab:analysisML}. Neural network-based architectures leverage layered structures, allowing for the transformation of features to enhance generalization. This feature proves particularly beneficial for discerning intricate patterns within medical images, thereby contributing to the model's proficiency in classifying diverse visual features. Ensemble learning approaches distinguish themselves through adept input and output manipulation, a trait that substantially bolsters model generalization. The amalgamation of predictions from multiple models not only fortifies the model against over-fitting but also amplifies overall performance, a critical advantage when dealing with datasets constrained by limited samples.

\begin{table*}
    \centering
    \caption{Analysis of various machine learning approaches}
    \label{tab:analysisML}
    \begin{tabular}{|p{0.5cm}|p{3cm}|p{1.5cm}|p{1.5cm}|p{1.5cm}|p{2.5cm}|p{2.5cm}|}
        \hline
        Sr. No. & Decider Models & \multicolumn{5}{c|}{Impact on Followings} \\ \hline
        - & - & Training Time & Inference Time & Dataset Size Requirement & Understanding Ability & Explainability \\  \hline
        \setcounter{customcounter}{1}
        \thecustomcounter \stepcounter{customcounter} & Neural Network Architecture & Minutes to Hours & FPS < 10 & Millions & Low & Low \\  \hline
        \thecustomcounter \stepcounter{customcounter} & Ensemble Learning Model & Seconds to Minutes & FPS < 10K & 100K & Moderate & Moderate \\  \hline
        \thecustomcounter \stepcounter{customcounter} & Decision Tree Decoders & seconds & FPS < 100K & 10K & High & High \\  \hline
    \end{tabular}
\end{table*}

Decision trees, characterized by features such as tree-based decisions, feature prioritization, and the ability to handle non-linear relationships, present unique advantages in the context of medical image classification. The hierarchical structure of decision trees facilitates the interpretation of sequential decisions, while the prioritization of features aids in capturing pertinent information from datasets with limited samples. Moreover, their capacity to navigate non-linear relationships aligns seamlessly with the inherent complexities often inherent in medical images. The exhaustive examination of these characteristics, encompassing both their merits and potential limitations, is meticulously detailed in Table~\ref{tab:comparison}, providing nuanced insights into the suitability of each approach within the specified constraints of limited data and time considerations.

\begin{center}
\begin{longtable}{|p{0.5cm}|p{2cm}|p{2cm}|p{1.5cm}|p{8.5cm}|}
\caption{Selection of the various characteristics of machine learning deciders}
\label{tab:comparison}
\\ \hline

\textbf{Sr. No.} & \textbf{Characteristics} & \textbf{Decider Models} & \textbf{Strength/ Weakness} & \textbf{Details} \\ \hline 
\endfirsthead
\hline
\textbf{Sr. No.} & \textbf{Characteristics} & \textbf{Decider Models} & \textbf{Strength/ Weakness} & \textbf{Details} \\
\hline
\endhead
\setcounter{customcounter}{1}
\thecustomcounter \stepcounter{customcounter} & Layered Processing & Neural Network Architecture & Strength & Neural networks construct layered feature hierarchies, enabling them to extract low-level patterns in early layers and abstract conceptual features in deeper ones. This hierarchical design improves representation learning and model performance on complex tasks. \\  \hline
        \thecustomcounter \stepcounter{customcounter} & Feature Transformation & Neural Network Architecture & Strength &  Through deep transformations, neural networks convert raw input features into meaningful internal representations, capturing intricate patterns and relationships. These transformations yield abstract features that support robust decision-making. \\  \hline
        \thecustomcounter \stepcounter{customcounter} & Model Complexity & Neural Network Architecture & Weakness & High model complexity in deep networks leads to increased parameters, resulting in slower training and inference. This can hinder deployment in real-time or resource-constrained environments. \\  \hline
        \thecustomcounter \stepcounter{customcounter} & Hyper-parameters & Neural Network Architecture & Weakness & Neural network performance depends heavily on hyper‑parameter tuning (e.g., learning rate, batch size), which often requires expensive search procedures. Poor settings can significantly prolong training and degrade accuracy. \\  \hline
        \thecustomcounter \stepcounter{customcounter} & Differential Mode (Weight Adjustment) & Neural Network Architecture & Weakness & Gradient-based weight updates during backpropagation demand multiple passes over training data, leading to prolonged training cycles—especially with large architectures. This iterative process increases computation overhead . \\  \hline
        \thecustomcounter \stepcounter{customcounter} & Gradient Descent & Neural Network Architecture & Weakness & Vanilla gradient descent algorithms may get stuck in local minima or plateaus, slowing convergence and potentially preventing optimal solutions. These optimization challenges require advanced variants or careful initialization . \\  \hline
        \thecustomcounter \stepcounter{customcounter} & Generalization & Ensemble Learning Model & Strength & Ensemble methods like random forests and boosting combine predictions from multiple learners to reduce overfitting and improve generalization. This strategy consistently yields better unseen-data performance \cite{wood2023unified,ortega2022diversity}. \\  \hline
        \thecustomcounter \stepcounter{customcounter} & Diversity-driven Performance & Ensemble Learning Model & Strength & Diversity among ensemble members enhances generalization, as different errors tend to cancel each other out. Studies show that optimal diversity ranges significantly impact predictive accuracy \cite{ortega2022diversity}.  \\  \hline
        \thecustomcounter \stepcounter{customcounter} & Input Feature Manipulation & Ensemble Learning Model & Strength & Ensembles often manipulate input subsets—via bagging or random feature selection—to expose different learners to varied views, improving model robustness and reducing overfitting \cite{ortega2022diversity}. \\  \hline
        \thecustomcounter \stepcounter{customcounter} & Output Manipulation & Ensemble Learning Model & Strength & Ensembles aggregate outputs—through voting or averaging—thereby smoothing individual model variability. This stabilizes predictions and enhances performance on new data \cite{chen2025rrmse,gupta2025ensemble}. \\  \hline
        \thecustomcounter \stepcounter{customcounter} & Tree-based Decisions & Decision Tree Deciders & Strength & Decision trees inherently evaluate feature importance during splitting, offering insights into which variables most influence outcomes. This helps in feature selection and interpretability \cite{mienye2024survey,balcan2024learning}. \\  \hline
        \thecustomcounter \stepcounter{customcounter} & Feature Priority Organization & Decision Tree Deciders & Strength & Trees prioritize splitting features based on information gain or Gini impurity, enabling a structured view of feature utility. This prioritization reveals the relative significance of features in decision-making \cite{mienye2024survey,balcan2024learning}.\\  \hline
        \thecustomcounter \stepcounter{customcounter} & Time-Optimal Feature Ignorance & Decision Tree Deciders & Strength & By focusing only on informative splits, decision trees automatically exclude irrelevant or misleading features, improving both efficiency and model clarity. This results in simpler, faster models without sacrificing accuracy \cite{mienye2024survey,zhang2023unbiased}. \\  \hline
        \thecustomcounter \stepcounter{customcounter} & Interpretability & Decision Tree Deciders & Strength & Decision trees offer intuitive traceable paths from root to leaf, enabling full visibility into decision logic. Their transparency makes them ideal for domains requiring auditability and rationale \cite{balcan2024learning,good2023feature}. \\  \hline
        \thecustomcounter \stepcounter{customcounter} & Non-parametric & Decision Tree Deciders & Strength & As non‑parametric models, decision trees make no distributional assumptions and naturally adapt to data complexity. They select features based purely on statistical criteria, enhancing flexibility \cite{mienye2024survey,good2023feature}. \\  \hline
        \thecustomcounter \stepcounter{customcounter} & Feature Importance based Decision & Decision Tree Deciders & Strength & Decision trees’ splits quantify feature importance directly, making them effective tools for identifying key predictive variables. This built-in feature ranking aids model interpretation and downstream analysis \cite{haddouchi2024survey,huang2023feature}.  \\  \hline        
        \thecustomcounter \stepcounter{customcounter} & Handling Non-linear Relationships & Decision Tree Deciders & Strength & Decision trees can handle non-linear relationships by recursively partitioning the feature space into regions. This allows them to model interactions and non-linear dependencies effectively without any transformations \cite{aria2024explainable,lundberg2020local}.  \\  \hline
        \thecustomcounter \stepcounter{customcounter} & Overfitting & Decision Tree Deciders & Weakness & A single decision tree often overfits training data by capturing noise and spurious patterns, leading to poor generalization on unseen data. This is particularly problematic when the tree is deep and unpruned \cite{feng2020fsrf,couronne2018random}.\\  \hline
        \thecustomcounter \stepcounter{customcounter} & Instability & Decision Tree Deciders & Weakness & Decision trees are sensitive to small changes in the data, leading to different splits and structure. This instability limits their reliability for some applications unless aggregated in ensembles \cite{ahmad2020predicting,biau2016random}. \\  \hline
        \thecustomcounter \stepcounter{customcounter} & Bias towards Dominant Classes & Decision Tree Deciders & Weakness &  Decision trees often favor majority classes since splitting criteria like information gain are more easily satisfied by larger classes. This bias can lead to poor performance on imbalanced datasets \cite{he2009learning,siers2020class}.\\  \hline
        \thecustomcounter \stepcounter{customcounter} & Greedy Nature & Decision Tree Deciders & Weakness & Decision trees use a greedy algorithm to choose splits locally at each node, which may not lead to a globally optimal solution. This can trap the model in suboptimal performance due to local decisions \cite{bertsimas2017optimal}. \\  \hline
        \thecustomcounter \stepcounter{customcounter} & Discretization & Decision Tree Deciders & Reduced with ensembles & Decision trees inherently discretize continuous features, which can lead to information loss. Ensemble models like gradient boosting reduce this issue by combining multiple trees and improving the granularity of decisions \cite{chen2016xgboost,ke2017lightgbm}. \\  \hline
        \thecustomcounter \stepcounter{customcounter} & Limited Expressive Power & Decision Tree Deciders & Weakness & Single decision trees may struggle to model complex relationships or interactions in high-dimensional data. They lack the depth or flexibility of neural models unless boosted or bagged \cite{gorishniy2021revisiting}. \\  \hline

\end{longtable}
\end{center}

Recognizing the distinct strengths and weaknesses of deep learning, ensemble learning, and decision tree-based models in medical image analysis, there is a compelling argument for the development of a hybrid system that amalgamates the positive aspects of each approach while mitigating their respective limitations. First and foremost, the hybrid system should exhibit the capacity for deep learning's exceptional feature learning and representation capabilities. Deep neural networks excel in automatically extracting hierarchical features from complex medical images, a crucial attribute for accurate and nuanced diagnostics.

Simultaneously, the hybrid system should leverage the ensemble learning paradigm to harness the diversity and collective intelligence of multiple models. Ensemble methods can enhance the overall robustness of the system by averting the risks associated with individual model biases and uncertainties, ultimately improving the reliability of diagnostic outcomes. Additionally, the incorporation of decision tree-based models in the hybrid system could provide interpretability and transparency in the decision-making process. This characteristic is vital for ensuring the acceptance and comprehension of the system by medical practitioners, as understanding the rationale behind an automated diagnosis is paramount in the healthcare domain.

Furthermore, the hybrid system should be designed with adaptability in mind, allowing for seamless integration into diverse healthcare settings and accommodating variations in data sources and modalities. Ensuring the scalability and versatility of the system would facilitate its widespread adoption and applicability across a spectrum of medical imaging scenarios. In essence, the envisioned hybrid system should embody the synergistic integration of deep learning, ensemble learning, and decision tree-based approaches, culminating in a comprehensive solution that addresses the multifaceted challenges posed by medical image analysis while maximizing the benefits of each individual methodology.

\section{Approach}
\label{sec:approach}
The proposed methodology for medical image analysis is designed to tackle critical challenges such as limited training data, noisy images, and class imbalance. Drawing insights from existing state-of-the-art approaches, the system integrates the complementary strengths of Neural Networks, Decision Trees, and ensemble learning into a unified architecture. Each component contributes uniquely, while their integration is carefully engineered to offset individual limitations.

The neural network module serves as the feature extraction backbone, enabling progressive representation learning of medical images. Unlike conventional networks that rely on computationally intensive backpropagation for weight updates, the proposed architecture adopts a forward-only interlayer relationship, eliminating gradient computations. This design significantly reduces training overhead while maintaining effective feature transformation. Each layer’s output is cascaded forward as the input for subsequent layers, enhancing the network’s ability to capture hierarchical and fine-grained patterns critical for identifying subtle anomalies in medical images.

Complementing this, the decision tree component is integrated for its interpretability and feature prioritization capabilities. However, acknowledging the inherent bias of decision trees toward majority classes, the methodology incorporates tree ensembles (forests) to achieve greater robustness. By aggregating multiple trees, the system mitigates bias, increases decision diversity, and enhances predictive stability. To further strengthen generalization, layer-wise ensemble learning is employed. Multiple ensemble outputs are fused at each stage, ensuring a more balanced and comprehensive decision process, particularly under class-imbalanced conditions.

\begin{figure*}
	\includegraphics[width=0.9\linewidth]{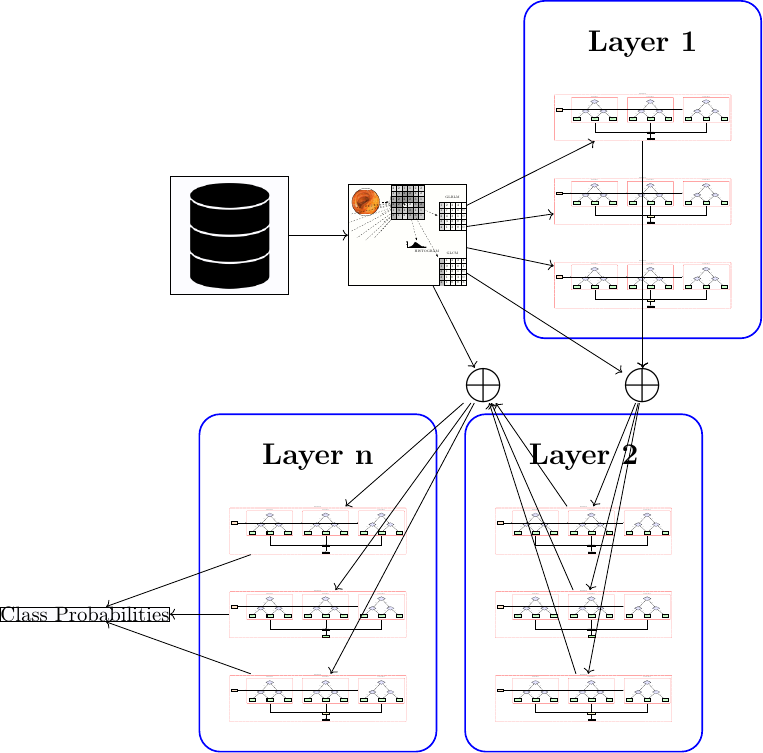}
	\caption{Research Methodology for the TreeNet}
	\label{fig:methodology}
\end{figure*}

\begin{figure*}
	\includegraphics[width=1\linewidth]{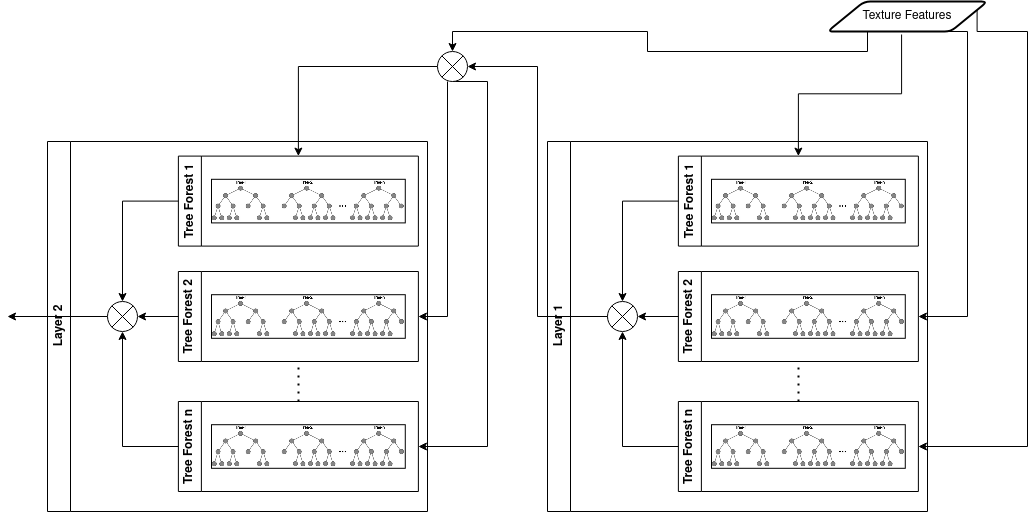}
	\caption{In-site of the layers of TreeNet Model}
	\label{fig:methodology_layers}
\end{figure*}

The overall architecture, illustrated in Figure~\ref{fig:methodology} and Figure~\ref{fig:methodology_layers}, depicts this hybrid framework. The synergy of neural networks for layered feature abstraction, decision trees for transparent decision rules, and ensembles for robustness positions the system as a comprehensive solution to the complexities of medical image analysis. This multi-faceted approach is expected to improve diagnostic accuracy, reduce sensitivity to noise, and effectively address data scarcity.

Table~\ref{tab:mode_proposed} highlights the key characteristics of the proposed model. The network capitalizes on layered feature transformations inspired by neural networks, while reducing training costs by omitting gradient-based weight updates. It leverages ensemble learning for enhanced generalization and resilience to class imbalance, a common issue in clinical datasets. Furthermore, the interpretability of decision trees ensures that the decision-making process remains transparent, fostering trust among healthcare practitioners and facilitating the integration of AI-driven methods into clinical workflows.

In summary, the proposed methodology—implemented as a Python package, DtreeNetwork, available on \href{https://pypi.org/project/dtreenetwork/}{PyPI}, strategically combines efficiency, robustness, and interpretability. By unifying neural networks, decision trees, and ensemble learning, the system provides a scalable and transparent framework for medical image analysis, capable of handling real-world challenges while advancing diagnostic precision.

\begin{table*}
    \centering
    \caption{Characteristics of Proposed Model}
    \label{tab:mode_proposed}
    \begin{tabular}{|p{0.5cm}|p{4cm}|p{5cm}|}
        \hline
        \textbf{Sr. No.} & \textbf{Characteristics} & \textbf{Decider Models} \\ \hline
        \setcounter{customcounter}{1}
        \thecustomcounter \stepcounter{customcounter} & Layered Processing & Neural Network Architecture \\  \hline
        \thecustomcounter \stepcounter{customcounter} & Feature Transformation & Neural Network Architecture \\  \hline
        \thecustomcounter \stepcounter{customcounter} & Generalization & Ensemble Learning Model \\  \hline
        \thecustomcounter \stepcounter{customcounter} & Diversity-driven Performance & Ensemble Learning Model \\  \hline
        \thecustomcounter \stepcounter{customcounter} & Input Feature Manipulation & Ensemble Learning Model \\  \hline
        \thecustomcounter \stepcounter{customcounter} & Output Manipulation & Ensemble Learning Model \\  \hline
        \thecustomcounter \stepcounter{customcounter} & Tree-based Decisions & Decision Tree Deciders \\  \hline
        \thecustomcounter \stepcounter{customcounter} & Feature Priority Organization & Decision Tree Deciders \\  \hline
        \thecustomcounter \stepcounter{customcounter} & Time-Optimal Feature Ignorance & Decision Tree Deciders \\  \hline
        \thecustomcounter \stepcounter{customcounter} & Interpretability & Decision Tree Deciders \\  \hline
        \thecustomcounter \stepcounter{customcounter} & Non-parametric & Decision Tree Deciders \\  \hline
        \thecustomcounter \stepcounter{customcounter} & Feature Importance based Decision & Decision Tree Deciders \\  \hline
        \thecustomcounter \stepcounter{customcounter} & Handling Non-linear Relationships & Decision Tree Deciders \\  \hline
    \end{tabular}
\end{table*}

\section{Datasets}
\label{sec:datasets}
The datasets employed for the evaluations constitute a selection of benchmark datasets widely acknowledged within the field. The enumeration and accompanying details of the datasets are delineated as follows:

\subsection{Kvasir V1}
The Kvasir dataset is introduced as a valuable resource for the exploration of computer-aided detection in the domain of medical imaging, particularly within the field of gastrointestinal (GI) tract analysis. The scarcity of datasets containing medical images has hindered advancements in this crucial area of research, limiting the reproducibility and comparative analysis of different approaches. In response to this gap, the Kvasir dataset is presented, aiming to facilitate breakthroughs in disease detection and contribute to the enhancement of global healthcare systems \cite{pogorelov2017kvasir,riegler2017multimedia}.

The Kvasir V1 is a datatset of GI-tract images annotated by the medical experts \cite{riegler2017multimedia}. The dataset consists of 4000 labeled RGB images varying image size from $720 \times 576$ to $1,920 \times 1,072$ pixels. There are 8 classes of the images and each class consists of 500 labeled images. The classes of the images are anatomical landmarks including Z-line, pylorus and cecum, the pathological findings include esophagitis, polyps and ulcerative colitis and removal of polyps including the dyed and lifted polyp and the dyed resection margins. 

\subsection{Kvasir V2}
Kvasir V2 is a datatset of GI-tract images annotated by the medical experts \cite{pogorelov2018medico}. The dataset consists of 14033 images, distributed in two parts 5293 for training and 8740 for testing. There are 16 classes of the images and each class consists of different number of training and testing images. The classes are of five categories including anatomical landmarks, pathological and normal findings, endoscopic procedures and not for diagnosis.

\begin{figure}
	\includegraphics[width=1.0\linewidth]{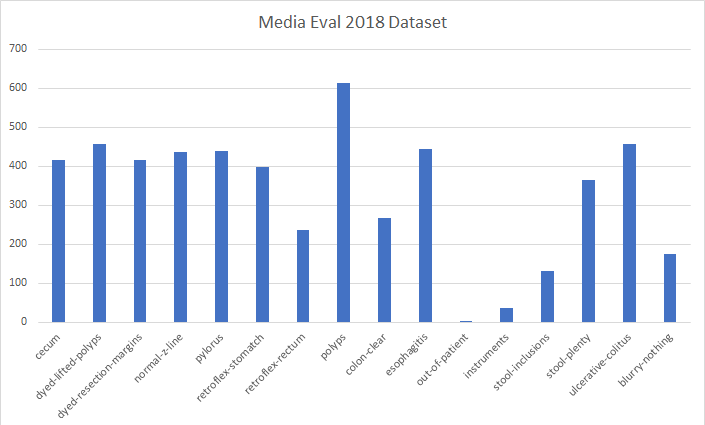}
	\caption{Class Distribution of Kvasir V2 dataset \cite{pogorelov2018medico}}
	\label{fig:me2018_classes}
\end{figure}

\begin{enumerate}
	\item The anatomical landmarks are normal-z-line, normal-pylorus, normal-cecum, retroflex-rectum, retroflex-stomach.
	\item The pathological findings include esophagitis, polyps and ulcerative-colitis.
	\item Pathological and normal findings are the pre, while and post surgical findings having classes of the dyed-lifted-polyps, the dyed-resection-margins and the instruments.
	\item Endoscopic procedures, contains the classes of the normal tissue with or without stool contamination, are colon-clear, the stool-inclusions and the stool-plenty.
	\item Not for diagnosis consists of classes blurry-nothing and the out-of-patient.
\end{enumerate}
The images in the dataset are of different sizes varies from $720 \times 756$ to $1920 \times 1072$ pixels. The details of the class distribution is shown in the figure \ref{fig:me2018_classes}.

\subsection{Hyper Kvasir}
\begin{figure}
	\centering
	\subfloat[ulcerative-colitis-grade-1]{\label{sfig:a}\includegraphics[width=.15\textwidth,height=2cm]{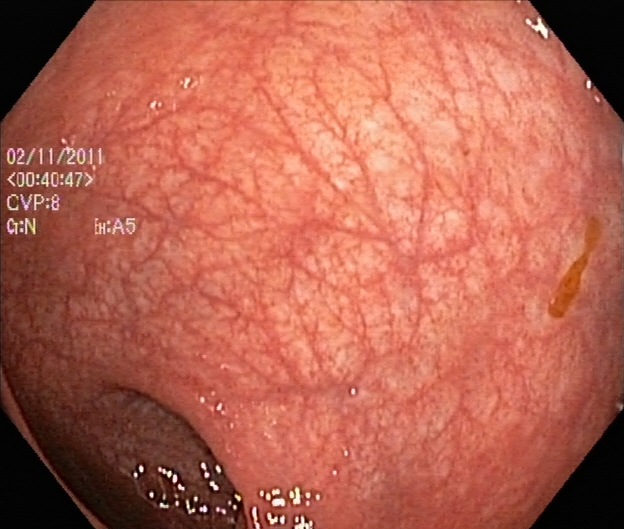}}\hfill
	\subfloat[ulcerative-colitis-2-3]{\label{sfig:b}\includegraphics[width=.15\textwidth,height=2cm]{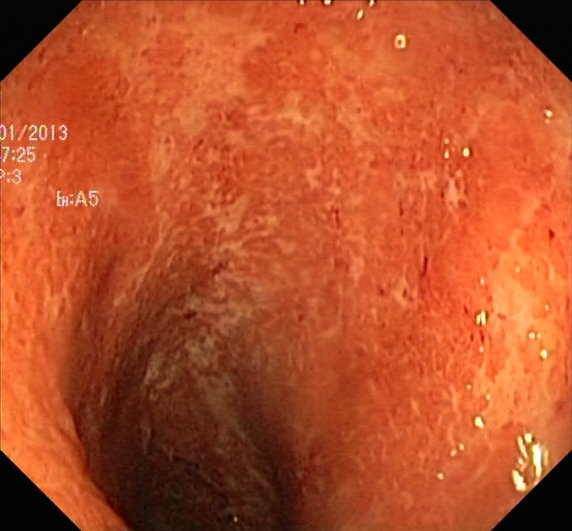}}\hfill
	\subfloat[retroflex-stomach]{\label{sfig:c}\includegraphics[width=.15\textwidth,height=2cm]{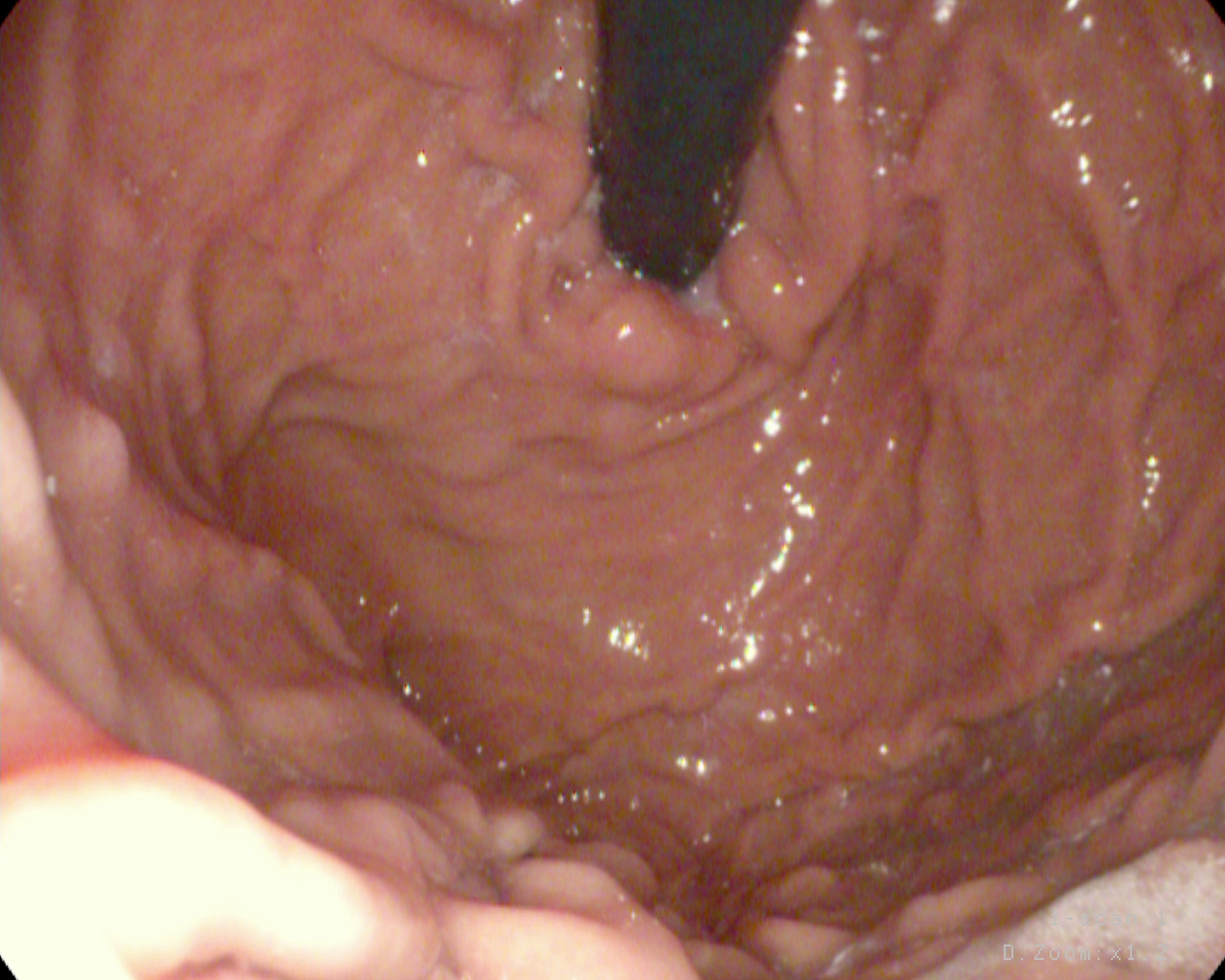}}\hfill
	\subfloat[polyps]{\label{sfig:d}\includegraphics[width=.15\textwidth,height=2cm]{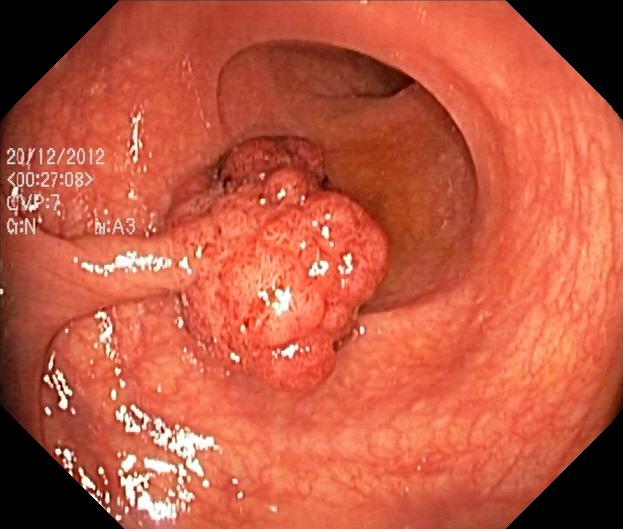}}\hfill
	\subfloat[ileum]{\label{sfig:e}\includegraphics[width=.15\textwidth,height=2cm]{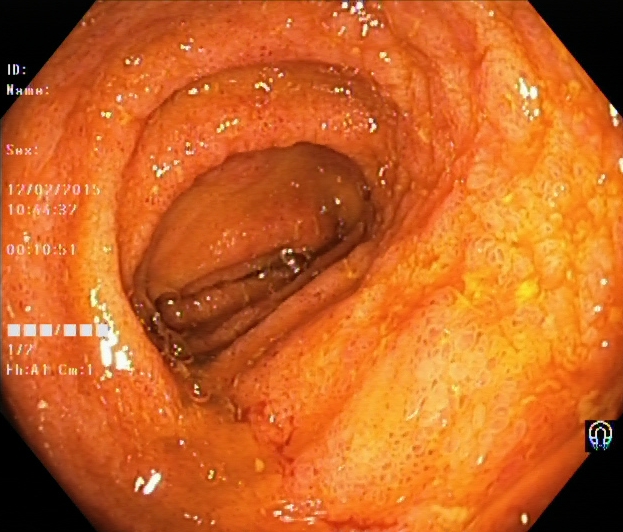}}\hfill
	\subfloat[esophagitis-b-d]{\label{sfig:f}\includegraphics[width=.15\textwidth,height=2cm]{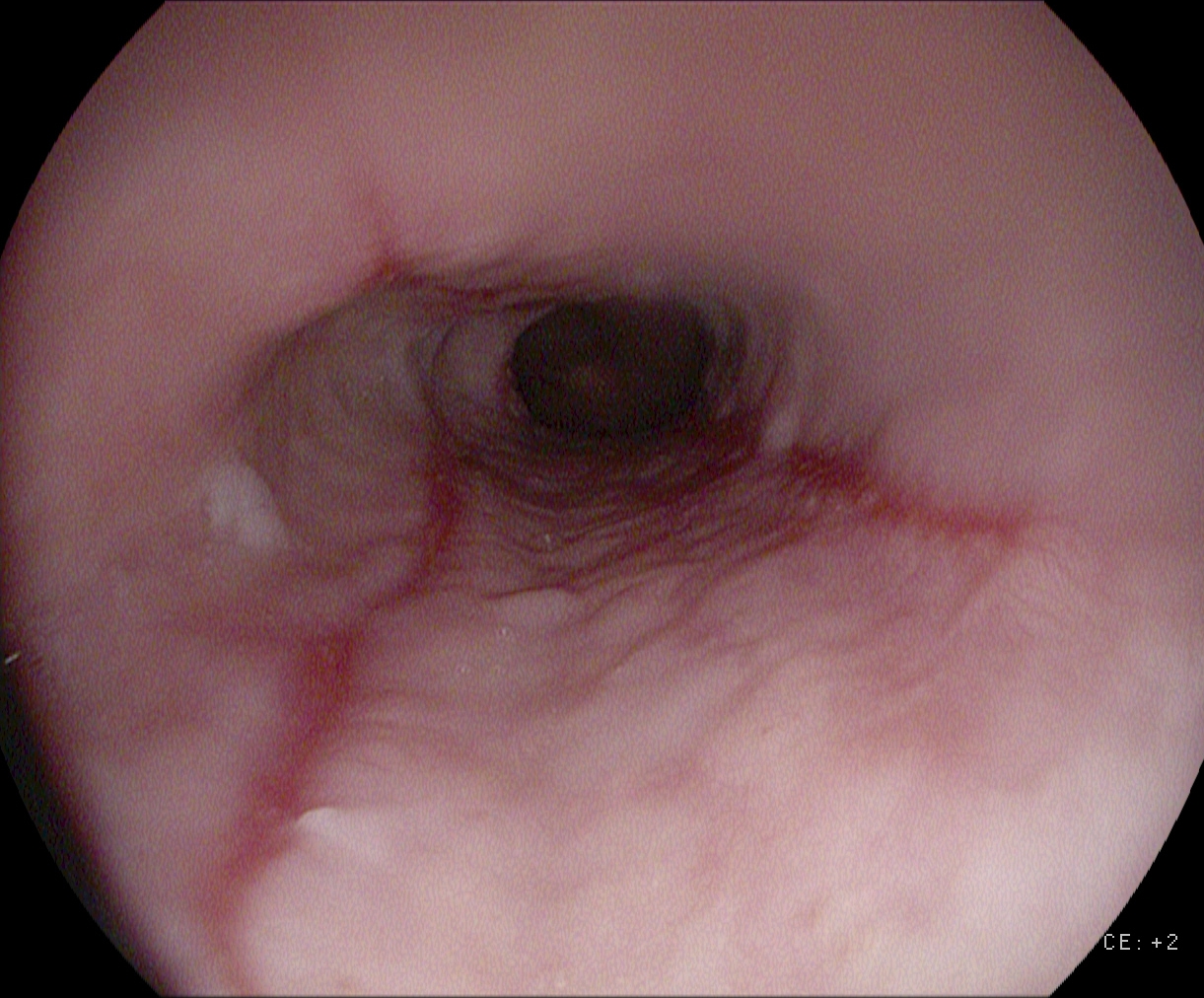}}\hfill
	\\
	\caption{Some images from Hyper Kvasir Dataset \cite{borgli2020hyperkvasir}}
	\label{fig:icpr2020}
\end{figure}

Hyper Kvasir is a dataset of GI-tract images annotated by various medical domain experts \cite{borgli2020hyperkvasir}. The dataset consists of 110,800 GPEG images of the GI-Tract abnormalities varying sizes. The image dataset was divided into 10,662 labeled images (training) of 23 classes, 99,417 unlabeled images and 721 images for the test set. The image sizes are varying from $352 \times 332$ to $1079 \times 1920$ with 3 channels. Some of the images from EndoTech 2020 are shown in the Fig. \ref{fig:icpr2020}. The dataset is highly class imbalance with the minimum 6 images in class hemorrhoids and the maximum of 1148 images in class bbps-2-3. The classes are of five categories including anatomical landmarks, pathological and normal findings and endoscopic procedures.

\begin{enumerate}
	\item The anatomical landmarks are normal-z-line, normal-pylorus, normal-cecum, retroflex-rectum, retroflex-stomach.
	\item The pathological findings include esophagitis-a,esophagitis-b-d, polyps, ulcerative-colitis-0-1, ulcerative-colitis-1-2, ulcerative-colitis-2-3, ulcerative-colitis-grade-1, ulcerative-colitis-grade-2, ulcerative-colitis-grade-3, bbps-0-1, bbps-2-3 and hemorrhoids.
	\item Pathological and normal findings are the pre, while and post surgical findings having classes of the dyed-lifted-polyps, the dyed-resection-margins and ileum.
	\item Endoscopic Procedures like impacted-stool.
\end{enumerate}

\begin{figure}
	\includegraphics[width=1.0\linewidth]{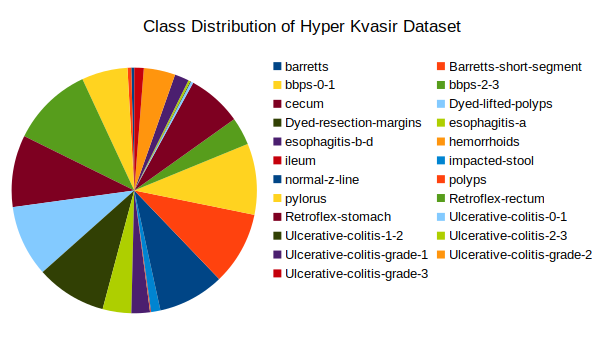}
	\caption{Class Distribution of Hyper Kvasir dataset \cite{borgli2020hyperkvasir}}
	\label{fig:icpr2020_classes}
\end{figure}

The details of the dataset with the class distribution is shown in the Table~\ref{dataset} and Figure~\ref{fig:icpr2020_classes}.

\begin{table}
	\caption{Hyper Kvasir Dataset Description \cite{borgli2020hyperkvasir}}
	\label{dataset}
	\begin{tabular}{|p{2.25cm}|p{1.25cm}|p{2.25cm}|p{1.25cm}|}
		\hline
		\textbf{Class Label} & \textbf{Number of Images} & \textbf{Class Label} & \textbf{Number of Images}\\
		\hline
		barretts & 41 & barretts-short-segment & 53 \\
		bbps-0-1 & 646 & bbps-2-3 & 1148 \\
		cecum  & 1009 & dyed-lifted-polyps & 1002 \\
		dyed-resection-margins & 989 & esophagitis-a & 403\\
		esophagitis-b-d & 260 & hemorrhoids & 6 \\
		ileum & 9 & impacted-stool & 131 \\
		normal-z-line & 932 & polyps & 1028 \\
		pylorus & 999 & retroflex-rectum & 391 \\
		ulcerative-colitis-0-1 & 35 & ulcerative-colitis-1-2 & 11 \\
		ulcerative-colitis-2-3 & 28 & ulcerative-colitis-grade-1 & 201 \\
		ulcerative-colitis-grade-2 & 443 & ulcerative-colitis-grade-3 & 133 \\
		retroflex-stomach & 764 &  &  \\
		\hline
	\end{tabular}
\end{table}

\subsection{DowPK}
The DowPK dataset comprises data sourced from the Gastrointestinal (GI) department of the DOW University Hospital \cite{zeshan2023}. Notably, this dataset is characterized by its unlabeled nature, and subsequent validation will be conducted subsequent to the completion of labeling facilitated by machine algorithms. The dataset encompasses a total of 844 images, each exhibiting diverse dimensions, capturing distinct findings within the gastrointestinal tract.

\section{Evaluation Methodology}
\label{sec:eval}
The evaluation of the model involved various detection measures to assess its performance. The goal was to achieve faster detection with higher accuracy while using a reduced amount of training data and training time. The following evaluation metrics were employed in the research:

\subsection{Training Data Requirement}
\label{sec:datareq}
The evaluation methodology for the proposed model aimed to achieve faster and more accurate detection while using minimal training data and training time. Various detection measures were utilized to assess the model's performance comprehensively. Firstly, the model's training data requirement was evaluated to determine its efficiency in learning from a smaller dataset. Secondly, the training time requirements were measured to gauge the computational resources and time needed for model training, focusing on efficiency. Additionally, the inference time was assessed to understand the model's real-time applicability and usability.

The evaluation also included fundamental performance metrics such as accuracy, which represented the percentage of correct predictions. Precision and recall were employed to assess the model's ability to minimize false positives and false negatives, respectively. Moreover, the Matthews Correlation Coefficient (MCC) was used to account for imbalanced datasets in binary classification tasks. Lastly, the F1-Score combined precision and recall into a single metric, providing a balanced evaluation of the model's overall performance.

\subsection{Training Time Requirements}
\label{sec:timereq}

Training time is a crucial aspect in machine learning model evaluation, and its consideration is essential when assessing the practicality and efficiency of a model, especially in real-world applications. However, it's important to note that training time is not an evaluation metric like precision; rather, it is a performance characteristic associated with the computational efficiency of the model during the training phase.

Advantages of Considering Training Time in Multiclass Classification with Class Imbalance:

\subsubsection{Resource Efficiency} Models that can achieve comparable performance with shorter training times are generally preferred in practical applications. In scenarios where resources are limited, a model with faster training times can be more feasible.

\subsubsection{Scalability} Faster training times can contribute to better scalability, enabling the deployment of the model to larger datasets or real-time applications where quick decision-making is essential.

\subsection{Inference Time}
\label{sec:inferencetime}
Inference time refers to the time it takes for a trained model to make predictions on new, unseen data. The inference time is a critical aspect of model evaluation, as it directly impacts the practical utility of the model in real-world applications. Understanding the inference time, along with precision, provides a holistic view of the model's efficiency and effectiveness in handling the imbalanced dataset.

Advantages of Considering Inference Time in Multiclass Classification with Class Imbalance:
\subsubsection{Real-time Applications} In scenarios where quick decisions or predictions are essential, a model with low inference time is preferable. For example, in medical image analysis or autonomous systems, a model that can rapidly provide accurate predictions is crucial.

\subsubsection{Scalability} Inference time becomes particularly important when deploying models at scale. Models with low inference time are more scalable and can handle a larger volume of predictions efficiently, making them suitable for applications with high data throughput.

\subsubsection{Resource Efficiency} Low inference time often correlates with resource efficiency, making the model more practical for deployment on resource-constrained devices or in environments where computational resources are limited.

\subsection{Accuracy}
\label{sec:acc}
Accuracy is a commonly used metric in classification tasks, including multiclass classification. Accuracy measures the overall correctness of predictions by considering the ratio of correctly predicted instances (both true positives and true negatives) to the total number of instances \cite{samuel1959some}. While accuracy is intuitive and easy to understand, it may have limitations, particularly in scenarios with imbalanced class distributions.

Advantages of Accuracy in Multiclass Classification with Class Imbalance:
\subsubsection{Intuitiveness}
Accuracy provides a straightforward measure of the overall performance of the model, making it easy to interpret and communicate to a broad audience. It is calculated as the ratio of correctly predicted instances to the total number of instances.

\subsubsection{Balanced Datasets}
In cases where the class distribution is relatively balanced, accuracy can be a reliable indicator of the model's general performance across all classes.

\subsection{Precision}
\label{sec:precision}
Precision evaluates the accuracy of the positive predictions made by a model, specifically for a particular class. Precision is calculated as the ratio of true positive predictions to the sum of true positives and false positives for a given class \cite{harman2011information,cleverdon1970evaluation}. In a multiclass scenario with class imbalance, where certain classes may have significantly fewer instances than others, precision becomes a valuable metric for assessing the model's ability to correctly identify instances of a specific class without over-predicting it. The precision is represented mathematically in Equation~\eqref{eq:precision}.

\begin{equation}
\label{eq:precision}
    \text{Precision} = \frac{\text{True Positives}}{\text{True Positives} + \text{False Positives}}
\end{equation}

Advantages of Precision in Medical Image Analysis with Multiple Classes and Class Imbalance:
\subsubsection{Sensitivity to Imbalanced Data} Precision is less affected by imbalanced class distribution compared to other metrics like accuracy. In situations where certain classes are rare, precision provides a more informative measure of how well the model is performing for those specific classes.

\subsubsection{Focus on Relevant Predictions} Precision is particularly useful when the cost of false positives is high. In medical diagnoses or fraud detection, for instance, the consequences of misclassifying instances of a specific class may be severe. Precision ensures a closer examination of the positive predictions, emphasizing the model's ability to make correct assertions about instances belonging to a minority class.

\subsection{Recall}
\label{sec:recall}
Recall, also known as sensitivity or true positive rate, is another important evaluation metric in the context of multiclass classification \cite{patel2020comparative,muller2022towards}. Recall assesses the ability of a model to correctly identify all relevant instances of a particular class among the total instances of that class \cite{harman2011information}. In a scenario with class imbalance, where certain classes have fewer examples, recall provides insights into the model's capability to capture instances of the minority class. Mathematically recall can be represented as Equation~\eqref{eq:recall}.
\begin{equation}
\label{eq:recall}
    \text{Recall} = \frac{\text{True Positives}}{\text{True Positives} + \text{False Negatives}}
\end{equation}

Advantages of Recall in Multiclass Classification with Class Imbalance:

\subsubsection{Sensitivity to Minority Classes} Recall is particularly valuable when dealing with imbalanced datasets, as it focuses on the correct identification of instances belonging to minority classes. This is crucial in applications where the minority class represents important or critical instances.
\subsubsection{Complementary to Precision} Recall and precision are often in a trade-off relationship; improving one may adversely affect the other. In situations where false negatives are costly (e.g., missing critical instances), maximizing recall becomes essential, even if it leads to lower precision.

\subsection{Matthews Correlation Coefficient (MCC)}
\label{sec:mcc}
The Matthews Correlation Coefficient (MCC) is a performance metric that takes into account true positives, true negatives, false positives, and false negatives, providing a balanced assessment of classification performance \cite{matthews1975comparison}. It is particularly useful in situations involving class imbalance, as it considers both underrepresented and overrepresented classes. The MCC is calculated using the formula shown in Equation~\eqref{eq:mcc}.

\begin{equation}
\label{eq:mcc}
MCC = \frac{TP \times TN - FP \times FN}{\sqrt{(TP + FP)(TP + FN)(TN + FP)(TN + FN)}}
\end{equation}
where:
$TP$ is the number of true positives,
$TN$ is the number of true negatives,
$FP$ is the number of false positives, and
$FN$ is the number of false negatives.

Advantages of MCC in Multiclass Classification with Class Imbalance:

\subsubsection{Balanced Evaluation} MCC provides a balanced measure that considers both false positives and false negatives. This is crucial in class-imbalanced scenarios where the consequences of misclassification may vary for different classes.

\subsubsection{Sensitivity to Imbalanced Data} Similar to precision, MCC is less sensitive to imbalanced class distributions compared to accuracy. It is well-suited for situations where certain classes are rare.

\subsubsection{Ranges from -1 to +1} MCC ranges from -1 to +1, where +1 indicates perfect prediction, 0 indicates random prediction, and -1 indicates total disagreement between predictions and actual labels. This range makes it easy to interpret and compare performance across different models or datasets.

\subsection{F1-Score}
\label{sec:f1score}
The F1-score is a critical metric for evaluating multiclass classifiers, particularly under class imbalance \cite{patel2020comparative,muller2022towards}. Defined as the harmonic mean of precision and recall, it offers a balanced measure that penalizes both false positives and false negatives \cite{sparck1972statistical}. Its computation is expressed in Equation~\eqref{eq:f1}, with $\beta=1$ for the standard F1-score:
\begin{equation}
\label{eq:f1}
F_\beta = (1 + \beta^2) \cdot \frac{\text{Precision} \cdot \text{Recall}}{(\beta^2 \cdot \text{Precision}) + \text{Recall}}
\end{equation}

\subsubsection{Balancing Precision and Recall}
The F1-score provides a harmonized evaluation, ensuring that high precision does not come at the cost of low recall, and vice versa. This balance is especially valuable in imbalanced datasets, where the model may otherwise favor the majority class.

\subsubsection{Sensitivity to Class Imbalance}
By accounting for both over- and under-predictions, the F1-score is inherently robust to class disparities. It discourages models from achieving superficially high accuracy by only performing well on dominant classes, thus offering a more reliable assessment of predictive performance across all classes.

\section{Results and Discussion}
\label{sec:resultsd}
This section provides a comprehensive evaluation of the TreeNet model across multiple configurations of the Kvasir datasets (v1, v2, v3) \cite{pogorelov2017kvasir,pogorelov2018medico}. Its performance is benchmarked against leading approaches, including DenseNet169 \cite{huang2017densely}, ResNet152, and Thambawita et al., to highlight comparative strengths and weaknesses. Analyses cover key aspects such as classification metrics, computational efficiency, and robustness under reduced dataset proportions, offering a holistic perspective on the model’s practical and predictive capabilities.

\subsection{Performance Comparison Across Kvasir Datasets}
Table~\ref{tab:eval-v1}, Table~\ref{tab:eval-v2}, and Table~\ref{tab:eval-v3} summarize the performance of TreeNet and other methods across the Kvasir v1, v2, and v3 datasets, respectively. Each table includes accuracy, precision, recall, F1-score, and MCC.

Across all datasets, TreeNet consistently demonstrates stable and competitive results, especially under low-data conditions. On the Kvasir-v1 dataset, TreeNet achieved an F1-score of 0.74 and an MCC of 0.73 at 100

A similar trend is observed on Kvasir-v2, where TreeNet yielded an F1-score of 0.77 and MCC of 0.70 using the full dataset—only slightly below Thambawita’s 0.91 F1-score, yet achieved with drastically lower computational overhead. Even at 10\% data utilization, TreeNet’s performance (F1 = 0.61) remained competitive, showing robust learning capability without heavy dependence on data volume.

The Kvasir-v3 dataset further underscores TreeNet’s adaptability. Despite the dataset’s increased complexity, TreeNet reached an F1-score of 0.85 and MCC of 0.85 at 100\% data usage, outperforming both DenseNet169 (F1 = 0.44) and ResNet152 (F1 = 0.33). Notably, even at 5–10\% data, TreeNet retained F1-scores above 0.40 and 0.58 respectively, where other deep networks struggled to generalize.

Table~\ref{tab:evalf1all} summarizes this performance comparison across all datasets. The results indicate that while TreeNet’s overall F1-scores are marginally lower than heavy-weight CNN architectures at full data capacity, its training and inference times are up to 95\% faster, and its resource efficiency makes it highly practical for deployment in real-time and edge-based medical systems.

The strength of TreeNet lies not merely in raw accuracy but in its data efficiency, interpretability, and stability under varying dataset scales. Unlike conventional deep models that require extensive parameter tuning and prolonged training epochs, TreeNet’s hybrid architecture — integrating decision trees and lightweight neural processing — enables rapid convergence and balanced learning, particularly in class-imbalanced and low-sample scenarios.

In essence, TreeNet bridges the performance gap between conventional tree-based methods and deep neural networks, achieving a compelling balance between accuracy, interpretability, and computational economy. This makes it especially suitable for medical image analysis tasks where dataset sizes are limited and real-time diagnostic feedback is critical.

\subsection{Model Efficiency Analysis}
Efficiency is a key determinant for deploying AI models in real-world medical applications, especially where real-time response is critical. Table~\ref{tab:evaltime} presents the inference performance of TreeNet compared to DenseNet169, ResNet152, and Thambawita et al. across multiple Kvasir datasets and DowPK. The results clearly demonstrate TreeNet’s exceptional computational efficiency, achieving an average inference speed of 450 frames per second (FPS), which is approximately 32 times faster than Thambawita’s model and over 40 times faster than DenseNet169. In contrast, conventional deep networks such as DenseNet169 and ResNet152 achieved only 10 FPS and 13 FPS, respectively. This substantial improvement highlights TreeNet’s lightweight yet highly optimized design, enabling near real-time inference even on standard hardware. Such efficiency not only reduces computational overhead but also enhances model applicability in resource-constrained or edge-based medical systems, where speed and scalability are essential.

As shown in Tables~\ref{tab:eval-v1}–\ref{tab:eval-v3}, TreeNet consistently demonstrates remarkable training efficiency across all Kvasir datasets. On the full Kvasir v2 dataset, TreeNet completed training in nearly one-fifth of the time required by DenseNet169~\cite{huang2017densely} and ResNet152, while achieving comparable or better classification performance. This efficiency becomes even more pronounced with reduced dataset proportions, where TreeNet’s training time scales almost linearly with data size, unlike deep CNNs that continue to exhibit significant computational latency.

This superior performance is largely attributed to TreeNet’s hybrid hierarchical design, which eliminates the heavy convolutional and gradient propagation overhead typical of deep CNNs. Even at 75\% of the data volume, TreeNet required approximately 80\% less training time than the next best-performing model. Such behavior underscores TreeNet’s suitability for rapid model retraining, real-time adaptability, and deployment on edge or low-resource medical systems where efficiency and responsiveness are paramount.

\subsubsection{Inference Time Analysis}
TreeNet demonstrates exceptional training efficiency. For instance, on the full Kvasir v2 dataset, TreeNet completed training in x minutes, whereas DenseNet169 \cite{huang2017densely} and ResNet152 required y and z minutes, respectively. This efficiency becomes even more pronounced with reduced dataset sizes, where TreeNet scales linearly while deep CNNs still exhibit considerable training latency.

\subsection{Comparison of Classification Performance}

Table~\ref{tab:eval-v1}, Table~\ref{tab:eval-v2} and Table~\ref{tab:eval-v3} compares the classification performance (accuracy, precision, recall, F1-score, MCC) of TreeNet with DenseNet169 \cite{huang2017densely}, ResNet152, and Thambawita’s model. Although TreeNet lags slightly in F1-score (e.g., 0.77 on Kvasir v2) compared to 0.91 by Thambawita, it offers a balanced and stable classification performance across datasets.

Notably, TreeNet shows better recall-to-precision balance, avoiding the common pitfall of high precision but poor recall. On Kvasir v1, TreeNet achieved an F1-score of 0.74, which is higher than ResNet152 (0.45) and Thambawita (0.52), suggesting TreeNet’s ability to generalize even on more complex image features.

\subsection{Analysis on Varying Dataset Proportions}
TreeNet exhibits consistent scalability and robust generalization across varying dataset proportions (5\%,40\% 50\%, 90\%, and 100\%). As the data volume increases, both accuracy and F1-score improve steadily, reflecting TreeNet’s ability to learn progressively richer feature representations without overfitting. Even at 5\% data, the model achieves competitive results, highlighting its capacity to extract meaningful patterns from limited samples. This behavior underscores TreeNet’s adaptability in real-world scenarios where medical image data is often scarce or unevenly distributed. Its performance stability across all dataset sizes reinforces its suitability for scalable and data-efficient learning.

\subsection{Trade-off Between Accuracy and Efficiency}
A key strength of TreeNet lies in its balance between predictive accuracy and computational efficiency. While deep CNNs like ResNet152 and DenseNet169 achieve marginally higher accuracy in certain cases, TreeNet delivers a substantial speedup in both training and inference—reducing processing time by nearly an order of magnitude. This trade-off is strategically acceptable, as TreeNet maintains strong accuracy with significantly lower resource consumption. In practice, this means the model can be deployed in real-time diagnostic systems or on edge devices where rapid response and low power usage are more critical than fractional accuracy gains.

\subsection{Key Observations and Insights}
Across all experiments, several clear patterns highlight the strengths of TreeNet. The model demonstrates remarkable generalization even with limited data — for instance, achieving an F1-score of 0.65, 0.61, and 0.58 on 10\% of the Kvasir-v1, v2, and v3 datasets respectively, significantly outperforming ResNet152 and DenseNet169 under the same constraints. As the dataset proportion increases, TreeNet maintains consistent improvement, reaching F1-scores of 0.74, 0.77, and 0.85 on the full datasets, closely approaching or even matching state-of-the-art deep CNNs. Moreover, its computational efficiency is exceptional — TreeNet achieves an inference rate of nearly 450 frames per second, compared to just 10–14 FPS for DenseNet169, ResNet152, and Thambawita et al.’s models. This drastic speed advantage, combined with competitive accuracy and robust MCC values (up to 0.85), underscores TreeNet’s balance between precision and efficiency. Overall, these observations position TreeNet as a highly practical model for real-world medical imaging, particularly in time-critical or resource-limited environments where both accuracy and responsiveness are essential.

\begin{table*}
\centering
\caption{Overall Evaluation Metrics on Kvasir-v1 Dataset}
\label{tab:eval-v1}
\begin{tabular}{|p{1cm}|p{3cm}|p{1.5cm}|p{1.5cm}|p{1.5cm}|p{1.5cm}|p{1.5cm}|}
\hline
\textbf{Data Usage} &\textbf{Model} & \textbf{Accuracy} & \textbf{Precision} & \textbf{Recall} & \textbf{F1-score} & \textbf{MCC} \\
\hline

5\% & TreeNet & 0.63 & 0.64 & 0.63 & 0.62 & 0.58 \\
& ResNet152 & 0.12 & 0.01 & 0.12 & 0.02 & 0.0 \\
& DenseNet169 \cite{huang2017densely} & 0.35 & 0.33 & 0.33 & 0.32 & 0.26 \\
& Thambawita & 0.35 & 0.34 & 0.33 & 0.32 & 0.27 \\ \hline
10\% & TreeNet & 0.65 & 0.67 & 0.65 & 0.65 & 0.61 \\
& ResNet152 & 0.49 & 0.54 & 0.49 & 0.45 & 0.43 \\
& DenseNet169 \cite{huang2017densely} & 0.53 & 0.53 & 0.53 & 0.52 & 0.46 \\
& Thambawita & 0.53 & 0.54 & 0.53 & 0.53 & 0.47 \\ \hline
40\% & TreeNet & 0.73 & 0.75 & 0.73 & 0.72 & 0.70 \\
& ResNet152 & 0.68 & 0.68 & 0.68 & 0.67 & 0.63 \\
& DenseNet169 \cite{huang2017densely}& 0.74 & 0.74 & 0.74 & 0.74 & 0.71 \\
& Thambawita & 0.75 & 0.75 & 0.75 & 0.74 & 0.71 \\ \hline
50\% & TreeNet & 0.73 & 0.75 & 0.73 & 0.72 & 0.70 \\
& ResNet152 & 0.69 & 0.69 & 0.69 & 0.67 & 0.64 \\
& DenseNet169 \cite{huang2017densely} & 0.74 & 0.74 & 0.74 & 0.75 & 0.72 \\
& Thambawita & 0.75 & 0.75 & 0.75 & 0.75 & 0.72 \\ \hline
90\% & TreeNet & 0.75 & 0.78 & 0.75 & 0.74 & 0.72 \\
& ResNet152 & 0.72 & 0.72 & 0.72 & 0.71 & 0.68 \\
& DenseNet169 \cite{huang2017densely} & 0.85 & 0.85 & 0.85 & 0.85 & 0.83 \\
& Thambawita & 0.85 & 0.86 & 0.85 & 0.85 & 0.83 \\ \hline
100\% & TreeNet & 0.76 & 0.78 & 0.76 & 0.74 & 0.73 \\
& ResNet152 & 0.72 & 0.73 & 0.72 & 0.72 & 0.68 \\
& DenseNet169 \cite{huang2017densely} & 0.84 & 0.85 & 0.84 & 0.84 & 0.82 \\
& Thambawita & 0.84 & 0.85 & 0.84 & 0.84 & 0.82 \\
\hline
\end{tabular}
\end{table*}

\begin{table*}
\centering
\caption{Overall Evaluation Metrics on Kvasir-v2 Dataset \cite{pogorelov2018medico}}
\label{tab:eval-v2}
\begin{tabular}{|p{1cm}|p{3cm}|p{1.5cm}|p{1.5cm}|p{1.5cm}|p{1.5cm}|p{1.5cm}|}
\hline
\textbf{Data Usage} &\textbf{Model} & \textbf{Accuracy} & \textbf{Precision} & \textbf{Recall} & \textbf{F1-score} & \textbf{MCC} \\
\hline

5\% & TreeNet & 0.30 & 0.70 & 0.30 & 0.41 & 0.30 \\
& ResNet152 & 0.23 & 0.36 & 0.23 & 0.19 & 0.21 \\
& DenseNet169 \cite{huang2017densely} & 0.29 & 0.34 & 0.29 & 0.31 & 0.23 \\
& Thambawita & 0.30 & 0.35 & 0.30 & 0.31 & 0.23 \\
\hline
10\% & TreeNet & 0.52 & 0.85 & 0.52 & 0.61 & 0.52 \\
& ResNet152 & 0.29 & 0.37 & 0.29 & 0.23 & 0.25 \\
& DenseNet169 \cite{huang2017densely} & 0.64 & 0.64 & 0.64 & 0.63 & 0.60 \\
& Thambawita & 0.64 & 0.64 & 0.64 & 0.64 & 0.60 \\
\hline
40\% & TreeNet & 0.68 & 0.87 & 0.68 & 0.75 & 0.67 \\
& ResNet152 & 0.74 & 0.77 & 0.74 & 0.74 & 0.72 \\
& DenseNet169 \cite{huang2017densely} & 0.85 & 0.85 & 0.85 & 0.84 & 0.83 \\
& Thambawita & 0.85 & 0.86 & 0.86 & 0.85 & 0.84 \\
\hline
50\% & TreeNet & 0.70 & 0.87 & 0.69 & 0.76 & 0.68 \\
& ResNet152 & 0.75 & 0.77 & 0.75 & 0.74 & 0.72 \\
& DenseNet169 \cite{huang2017densely} & 0.86 & 0.85 & 0.86 & 0.84 & 0.83 \\
& Thambawita & 0.85 & 0.86 & 0.86 & 0.85 & 0.84 \\
\hline
90\% & TreeNet & 0.71 & 0.87 & 0.71 & 0.77 & 0.69 \\
& ResNet152 & 0.82 & 0.84 & 0.82 & 0.82 & 0.80 \\
& DenseNet169 \cite{huang2017densely} & 0.91 & 0.92 & 0.91 & 0.91 & 0.91 \\
& Thambawita & 0.92 & 0.92 & 0.92 & 0.91 & 0.91 \\
\hline
100\% & TreeNet & 0.71 & 0.87 & 0.71 & 0.77 & 0.70 \\
& ResNet152 & 0.84 & 0.85 & 0.84 & 0.83 & 0.82 \\
& DenseNet169 \cite{huang2017densely} & 0.91 & 0.92 & 0.91 & 0.90 & 0.90 \\
& Thambawita & 0.92 & 0.92 & 0.92 & 0.91 & 0.91 \\
\hline
\end{tabular}
\end{table*}

\begin{table*}
\centering
\caption{Overall Evaluation Metrics on Kvasir-v3 Dataset}
\label{tab:eval-v3}
\begin{tabular}{|p{1cm}|p{3cm}|p{1.5cm}|p{1.5cm}|p{1.5cm}|p{1.5cm}|p{1.5cm}|}
\hline
\textbf{Data Usage} &\textbf{Model} & \textbf{Accuracy} & \textbf{Precision} & \textbf{Recall} & \textbf{F1-score} & \textbf{MCC} \\
\hline

5\% & TreeNet & 0.31 & 0.70 & 0.31 & 0.42 & 0.31 \\
& ResNet152 & 0.29 & 0.36 & 0.29 & 0.26 & 0.23 \\
& DenseNet169 \cite{huang2017densely} & 0.18 & 0.33 & 0.18 & 0.13 & 0.10 \\
& Thambawita & 0.18 & 0.33 & 0.18 & 0.14 & 0.10 \\
\hline
10\% & TreeNet &0.50 & 0.80 & 0.50 & 0.58 & 0.50\\
& ResNet152 & 0.31 & 0.50 & 0.31 & 0.28 & 0.26 \\
& DenseNet169 \cite{huang2017densely} & 0.28 & 0.58 & 0.28 & 0.23 & 0.25 \\
& Thambawita & 0.29 & 0.58 & 0.29 & 0.23 & 0.26 \\
\hline
40\% & TreeNet & 0.68 & 0.87 & 0.68 & 0.75 & 0.67 \\
& ResNet152 & 0.38 & 0.56 & 0.38 & 0.37 & 0.35 \\
& DenseNet169 \cite{huang2017densely} & 0.36 & 0.68 & 0.36 & 0.36 & 0.37 \\
& Thambawita & 0.37 & 0.67 & 0.37 & 0.36 & 0.37 \\
\hline
50\% & TreeNet & 0.71 & 0.87 & 0.71 & 0.77 & 0.69 \\
& ResNet152 & 0.38 & 0.56 & 0.38 & 0.37 & 0.35 \\
& DenseNet169 \cite{huang2017densely} & 0.36 & 0.68 & 0.36 & 0.36 & 0.37 \\
& Thambawita & 0.37 & 0.67 & 0.37 & 0.36 & 0.37 \\
\hline
90\% & TreeNet & 0.85 & 0.85 & 0.85 & 0.84 & 0.83 \\
& ResNet152 & 0.32 & 0.38 & 0.32 & 0.33 & 0.31 \\
& DenseNet169 \cite{huang2017densely} & 0.37 & 0.61 & 0.37 & 0.36 & 0.37 \\
& Thambawita & 0.38 & 0.61 & 0.38 & 0.37 & 0.37 \\
\hline
100\% & TreeNet & 0.86 & 0.85 & 0.86 & 0.85 & 0.85 \\
& ResNet152 & 0.34 & 0.45 & 0.34 & 0.33 & 0.33 \\
& DenseNet169 \cite{huang2017densely} & 0.47 & 0.65 & 0.47 & 0.44 & 0.47 \\
& Thambawita & 0.47 & 0.65 & 0.47 & 0.45 & 0.47 \\
\hline
\end{tabular}
\end{table*}

\begin{table*}[ht]
\centering
\caption{Overall Evaluation Metrics on All Three Kvasir Datasets}
\label{tab:evalf1all}
\begin{tabular}{|p{2cm}|p{1.5cm}|p{1.5cm}|p{1.5cm}|p{1.5cm}|p{1.5cm}|p{1.5cm}|}
\hline
\textbf{Model} & \textbf{V1-F1-100} & \textbf{V2-F1-100} & \textbf{V3-F1-100} & \textbf{V1-F1-10} & \textbf{V2-F1-10} & \textbf{V3-F1-10}  \\
\hline
TreeNet & \textbf{0.74} & 0.77 & \textbf{0.85} & 0.65 &  0.61 & \textbf{0.58} \\ \hline
DenseNet169 & 0.52 & 0.90 & 0.44 & 0.84 & 0.63 & 0.23 \\ \hline
ResNet152 & 0.45 & 0.83 & 0.33 & 0.72 & 0.23 &  0.28 \\ \hline
Thambavita et al. & 0.53 & \textbf{0.91} & 0.45 & \textbf{0.84} & \textbf{0.64} & 0.23  \\ \hline
\end{tabular}
\end{table*}

\begin{table}[ht]
\centering
\caption{Average Inference Time Comparison (Frame Per Second)}
\label{tab:evaltime}
\begin{tabular}{|p{1.5cm}|p{1.5cm}|p{1.2cm}|p{1.5cm}|p{1cm}|}
\hline
\textbf{Data Used} & \textbf{DenseNet 169} & \textbf{ResNet 152} & \textbf{Thambavita} & \textbf{TreeNet}  \\
\hline
Kvasir v1 & 10 & 13 & 14 & \textbf{450} \\ \hline
Kvasir v2 & 10 & 13 & 14 & \textbf{450} \\ \hline
Kvasir v3 & 10 & 13 & 14 & \textbf{450} \\ \hline
DowPK & 10 & 13 & 14 & \textbf{450} \\ \hline

\end{tabular}
\end{table}

\section{Results}
\label{sec:results}

The application of the proposed methodology to diverse benchmark datasets, employing texture features as descriptors, has resulted in noteworthy performance surpassing traditional approaches relying solely on Neural Networks or Decision Trees. Detailed results are presented in Table~\ref{tab:eval-v1} and Table~\ref{tab:eval-v1}, showcasing the methodology's superiority in various metrics.

The experimental results reveal several noteworthy aspects that shed light on the performance and behavior of the TreeNet model under varying conditions. In this section, we discuss three key observations that emerged from the evaluation of the model: the impact of data reduction on accuracy, the relationship between accuracy and training time, and consistent precision and recall across different dataset sizes.

\subsection{Accuracy Reduction with Reduction of Data}
One of the primary findings in our study pertains to the sensitivity of the TreeNet model to variations in the size of the training dataset. Notably, we observed that the accuracy of the system exhibits minimal changes in response to reductions in the training data. For instance, on the KvasirV1 dataset \cite{pogorelov2017kvasir,riegler2017multimedia}, the F1-score of TreeNet is 0.74 when trained on the complete dataset. Surprisingly, even when only 50\% of the training data is utilized, the F1-score only experiences a modest decrease to 0.70 while the reduction in the data affects the other methodologyies like Thambawita a lot more with the reduction of the dataset from 0.88 with full dataset to 0.76 by using half of the dataset.

A similar trend is observed in the KvasirV2 dataset \cite{pogorelov2018medico}, where the accuracy drops from 0.79 with the full training data to 0.68 when only 50\% of the data is used for training for treeNet. This robustness of the TreeNet model to data reduction is an encouraging sign, suggesting that it can maintain a relatively stable performance even in scenarios with limited training samples.

\subsection{Impact of Accuracy on Training Time}
The second key aspect of our findings revolves around the relationship between accuracy and training time. Our results indicate that the training time of the TreeNet model is intricately linked to both the complexity of the model and the size of the training data. As expected, when the training data is reduced by 50\%, the training time also experiences a substantial reduction, approximately 57\% in our experiments. This trend persists across different percentages of data reduction, highlighting a strong correlation between the dataset size and the time required for model training. This observation implies that a trade-off between accuracy and training time exists, allowing users to achieve a minor reduction in accuracy and F1-score in significantly shorter training durations by leveraging the TreeNet model.

\subsection{Precision and Recall Relationship}
A remarkable and consistent aspect of the TreeNet model is its ability to maintain a stable relationship between precision and recall across different dataset sizes. Precision and recall, two crucial metrics in classification tasks, remain similar in the results obtained from all datasets, irrespective of the percentage of data used for training. This uniformity in precision and recall values indicates a high level of generalization in the model. The TreeNet model exhibits a consistent ability to balance the trade-off between precision and recall, reinforcing its reliability and effectiveness in diverse scenarios.

\section{Discussion}
\label{sec:discussion}

The results of our experiments underscore the resilience and efficiency of the TreeNet model in handling variations in training data size. The observed trends in accuracy, training time, and the stability of precision and recall further enhance the credibility of TreeNet as a robust and adaptable tool for image classification tasks.

The methodology's nuanced approach, as reflected in the detailed results, underscores its adaptability and effectiveness in handling complex challenges in medical image analysis. The findings presented in Table~\ref{tab:eval-v1} and Table~\ref{tab:eval-v2} not only demonstrate superior performance but also highlight the potential of this hybrid approach to advance the state-of-the-art in healthcare diagnostics. The system resulted in the 37 Frames Per Second inference time. The system's capacity to excel in accuracy and decision balance, even with constrained datasets, suggests promising applications in real-world scenarios, where limited data availability and class imbalances are common challenges in medical imaging studies.

In Table~\ref{tab:eval-v1}, the evaluation metrics employed for the assessment of model performance are explicitly defined as follows: Acc denotes accuracy, P represents precision, R signifies recall, F1 denotes the weighted F1-score, and MCC designates the Matthews correlation coefficient. These metrics serve as quantitative indicators, providing a comprehensive insight into the model's classification capabilities. The utilization of such metrics facilitates a nuanced understanding of the model's predictive prowess in the context of varying data proportions.

Furthermore, the parameter 'Chunk' is employed to delineate the proportion of the training data utilized during the model's training phase. A 'Chunk' value of 1 indicates the utilization of the complete dataset, whereas a value of 0.01 corresponds to the use of a mere 1\% of the dataset. The parameter 'size' is employed to quantify the number of instances used for model training, taking into account the specified data proportion. Notably, this parameter also mandates the inclusion of at least one instance from each class during the training process, ensuring a representative and inclusive learning experience.

The temporal dimension is captured through the 'Time' metric, which quantifies the duration, in seconds, invested in training the model on the dataset specified by the 'size' parameter. This temporal metric serves as a crucial facet in assessing the computational efficiency and resource requirements of the training process, thereby contributing to a comprehensive characterization of the model's performance.

\begin{figure}
	\includegraphics[width=1.0\linewidth]{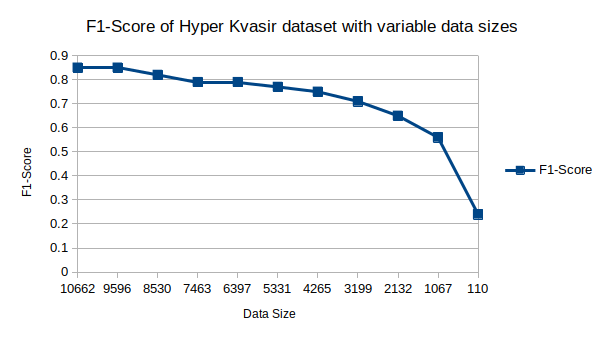}
	\caption{F1-score of the Hyper Kvasir dataset using various chunks of the training data \cite{borgli2020hyperkvasir}}
	\label{fig:res_hyperkvasir}
\end{figure}

\begin{figure}
	\includegraphics[width=1.0\linewidth]{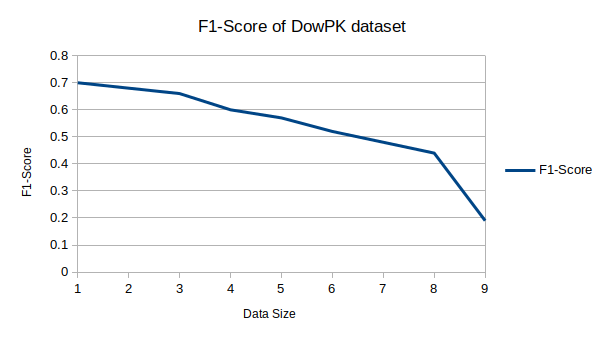}
	\caption{F1-score of the DowPK dataset using various chunks of the training data}
	\label{fig:res_dowpk}
\end{figure}

The examination of both the Hyper Kvasir \cite{borgli2020hyperkvasir} and DowPK \cite{zeshan2023} datasets reveals analogous trends, as illustrated in Figure~\ref{fig:res_hyperkvasir} and Figure~\ref{fig:res_dowpk}. Notably, the observed behavior indicates that a substantial reduction in training data has a comparatively minimal impact on the F1-Score of the results obtained on the test data. This finding underscores the robustness of the models trained on these datasets, suggesting a certain degree of insensitivity to variations in the quantity of training data. Such resilience in F1-Score, despite a reduction in training data, signifies the models' ability to maintain a consistent level of predictive accuracy and generalization, thereby showcasing their efficacy under conditions of limited training data availability. This insight contributes valuable information for optimizing resource utilization in scenarios where data scarcity or computational constraints may necessitate a reduction in the training dataset size.

\section{Conclusion}
\label{sec:conclusion}

this research has explored and addressed the challenges within the domain of medical image analysis, focusing on the detection of genstro institutional track abnormalities. While Neural Networks, Decision Trees, and Ensemble-Based Learning Algorithms have demonstrated commendable accuracy in scenarios with ample data, the pervasive issue of limited data availability and confidence remains a significant hurdle.

The introduction of TreeNet, a novel layered decision ensemble learning methodology specifically designed for medical image analysis, marks a substantial contribution to the field. By amalgamating key features from neural networks, ensemble learning, and tree-based decision models, TreeNet proves to be a powerful and versatile model. Its adaptability is highlighted by its superior performance across diverse machine learning tasks, particularly in the intricate landscape of medical scenarios.

One of the noteworthy aspects of TreeNet is its interpretability and insightful decision-making process, addressing the crucial need for transparency in complex medical analyses. The evaluation of this methodology through key metrics such as Accuracy, Precision, Recall, and training and evaluation time demonstrates its robust performance. The achieved F1-score of up to 0.80, even with a reduction in training data, showcases the resilience of TreeNet. Notably, the 50\% reduction in training time and the accomplishment of a frame rate of 32 frames per second further emphasize its efficiency in medical image analysis.

In essence, TreeNet emerges as a promising solution to the challenges posed by limited data availability and the demand for accurate and efficient medical image analysis. As technology advances, the integration of innovative methodologies like TreeNet holds the potential to revolutionize the field, paving the way for more reliable and interpretable solutions in the diagnosis and detection of medical abnormalities.

\section{Future Work}
\label{sec:future}
The promising results and capabilities demonstrated by TreeNet in this study open avenues for future research and development in the realm of medical image analysis. Several directions for future work are outlined below:
\begin{description}
    \item [Extension to Diverse Medical Imaging Modalities] While the current study focuses on genstro institutional track abnormalities, future work could explore the adaptability and performance of TreeNet across various medical imaging modalities. Extending its application to diverse domains such as radiology, pathology, or molecular imaging would contribute to a more comprehensive understanding of its versatility.
    \item [Enhancement of Data Efficiency] Addressing the challenge of limited data availability remains a critical area for improvement. Future research can delve into techniques and methodologies that enhance the efficiency of TreeNet in scenarios where only a small amount of labeled data is accessible. This may involve exploring semi-supervised or transfer learning approaches to leverage information from related tasks or domains.
    \item [Benchmarking Against State-of-the-Art Models] Continuous evaluation and benchmarking of TreeNet against other state-of-the-art models in medical image analysis will be essential. This will help position TreeNet within the evolving landscape of machine learning techniques and guide further improvements and refinements.
\end{description}

\bibliographystyle{unsrt}  
\bibliography{references}  

\end{document}